\documentclass[twocolumn ]{deepclass1}

\usepackage{graphicx}   % Including figure files
\usepackage{amsmath}    % Advanced maths commands
\usepackage{amssymb}    % Extra maths symbols
\usepackage{multirow}
\usepackage{float}
\usepackage{xcolor}
\usepackage{multirow}
\usepackage{natbib}
\newcounter{note}
\let\oldmarginpar\marginpar
\renewcommand\marginpar[1]{\-\oldmarginpar[\raggedleft\footnotesize #1]%
{\raggedright\footnotesize #1}}

\definecolor{Dred}{rgb}{0.312,0.070,0.070}

\shorttitle{Machine learning applied in the multi-scale 3D stress modelling}

\begin{document} 

\title{Machine learning applied in the multi-scale 3D stress modelling in geological structures}

\correspondingauthor{Xavier Garcia \& Adrian Rodriguez}
\email{xteijeiro@slb.com \&aherrera9@slb.com}

\author{Xavier Garcia Teijeiro}
\author{Adrian Rodriguez-Herrera}
\affiliation{Schlumberger Digital and Integration. London, UK}

\begin{abstract}

This paper proposes a methodology to estimate stress in the subsurface by a hybrid method combining finite element modeling and neural networks. This methodology exploits the idea of obtaining a \emph{multi-frequency solution} in the numerical modeling of systems whose behavior involves a wide span of length scales. One \emph{low-frequency} solution is obtained via inexpensive finite element modeling at a coarse scale. The second solution provides the fine-grained details introduced by the heterogeneity of the free parameters at the fine scale. This \emph{high-frequency} solution is estimated via neural networks pre-trained with partial solutions obtained in high-resolution finite-element models. When the coarse finite element solutions are combined with the neural network estimates, the results are within a 2\% error of the results that would be computed with high-resolution finite element models. This paper discusses the benefits and drawbacks of the method and illustrates their applicability via a worked example.

\end{abstract}


\section{Introduction}
\label{sec:intro}

In science and engineering the behavior of a variety of physical systems is often described by a set of equations. In such case, the standard approach in numerical modelling of those system is that of solving the appropriate discretized versions of partial differential equations by suitable numerical algorithms such as the finite difference method or the finite element method. 

In geotechnics, the de-facto approach to estimate stress and deformation in the subsurface is by solving the equations of elasticity theory via finite element modelling. Stress modelling can be challenging  when the behavior of the physical system studied involves a wide span of length and/or time scales. This is because partial differential equations need to be discretized to be solved numerically. In mesh-based methods, such as the finite element, the volume of interest needs to be divided in sub-domains or cells. The cell size is the minimum length scale at which a solution can be obtained locally; it is the resolution of the numerical solution. As the cell-count increases, so does the computational cost of solving the equations \citep{Efendiev2009}.
 
The cell count in a model can increase because the volume under study needs to be large in comparison to the cell size to model a specific physical phenomenon. Hence, more cells are needed to discretize the volume for a given mesh resolution. The cell count can also increase when a high resolution (fine-scale) is needed in a particular application. In this case, the cells need to be smaller and therefore more are needed to discretize the same volume. Multi-scale problems may require large models and high resolution at the same time. 

As an example, when prospecting for hydrocarbon reservoirs, the areas under exploration can extend hundreds of squared kilometers in the horizontal  plane and several kilometers in depth. Within this volume it is often needed to estimate the state of in-situ stress in thin zones of the stratigraphic column, or the stress contrast between thin adjacent layers. Common applications include the study of safe drilling pressure windows, the analysis of hazards when drilling nearby geological faults or the assessment of trapping mechanism during hydrocarbon migration. Estimating the in-situ stress is also important in hydraulic fracturing since high stress concentration in thin layers can act as a barrier for the vertical propagation of the fractures \citep{warpinski, garcia}.  The  horizontal propagation of the fractures is also influenced by stress heterogeneity but at a scale in the order of hundreds of meters. In yet a larger scale, the length of horizontal wells is in the range of kilometers. Hence, they sample a varying stress field along their trajectory  
\citep{Berard, Ueda}. At this scale, stress variations due geological structures and rock heterogeneity can affect well feasibility and performance.

The selection of sweet-spots and the assessment of the feasibility of hydrocarbon plays depend on these details at all different length scales.  To tackle these problems, both the size and the resolution of numerical models needs to be increased in order to capture the heterogeneity of the mechanical properties of the rock. Given the relatively large volumes of interest, the computational cost in this kind of problems can render these workflows impractical.

One alternative to cope with the computational overhead is to make use of upscaling techniques. These aim at solving the relevant equations in a coarse and manageable resolution. Previous to simulation,  the properties of the physical system are represented as  effective properties at the coarse scale while attempting to capture as much as possible the characteristic behavior of the fine scale. 
The use of upscaling techniques is common in geophysics and in applications of reservoir modelling for flow simulation 
%\citep{backus, menezes2003, chalon2004, wigmosta,florio}. 
\citep{backus, menezes2003, chalon2004, florio}. 
Note that depending on the model size and the problem studied, even the coarse scale solution may require the use of multi-scale approaches at the solver level \cite{Efendiev2009,buck,castelletto}.
 
Note that upscaling can reduce the cost of the simulations, but it is not an alternative when a high-resolution solution is needed. As an example, that is the case when estimating the effects of sub-seismic resolution uncertainty in the stress field \citep{trudeng}. 

An option in such cases would be to combine upscaling and downscaling techniques. Downscaling is the process of relocating coarse resolution into a fine spatial scale 
\citep{maraunwidmann2018, gaur}.
For instance, \cite{ita2015a}
developed a method to upscale mechanical properties and pressures to a coarse scale and obtained solutions for pore volume changes. The solutions were then downscaled to a high-resolution fluid flow simulator. See also \citep{Efendiev2009, Efendiev2013Book, babei, nunna2017a, brouwer2013a}.

Static (or out-of-solver) downscaling aims at inferring high-resolution information from the coarse scale without an explicit hierarchical mapping between the fine and coarse degrees of freedom. Instead,  the method relies in applying transformation rules between variables at different scales 
\citep{ren,qiu,torrealba2019a}.
One difficulty of statistical downscaling is that of finding suitable transformations between coarse and fine resolutions is not trivial. 

This paper explores the idea of applying physics-informed transformations that, when applied to the solutions of coarse finite element models, the results are similar to those that would be obtained by solving the corresponding partial differential equations (PDEs) with high spatial resolution. The methodology exploits in the idea of a “multi-scale solution”: one solution for the coarse scale provided by finite element modeling and another solution for a fine scale as estimated via neural networks previously trained with high-resolution finite element solutions. As such, the method proposed here can be considered a stress downscaling method based on machine learning techniques. While similar approaches have been  used to in other disciplines, see the review by \cite{xu} or the recent comparison of techniques in \cite{vandal} among others \citep{sailor2000,maraun2010,tarmizi}, this paper addresses stress downscaling applications in geosciences.

The paper is organized as follows: Section 2, will define the problem that we are aiming to solve. The strategy proposed will be discussed thereafter. Section 3  will provide details about the model that will be used throughout the paper in order to test the method. Section 4 will implement the strategy proposed. The last two sections, Section 5 and Section 6, will be dedicated to conclusions and potential future directions.

\section{	Problem statement}\label{problem}
Finite element models estimate the subsurface stress by solving constitutive equations relating the elastic properties of the rock, the pressure of the subsurface fluids and the tectonic forces acting. The linear-elastic model aims at solving the equation \citep{timosenko}:
\begin{equation}\label{eq1}
\sigma_{ij} = \kappa_{ijkp}\epsilon_{ij} -P_{p} + f
\end{equation}
under the equilibrium constraint:
\begin{equation}\label{equilibriumcdond}
\sigma_{ij,j} = +f_i  = 0. 
\end{equation}
Here it is used the shortened notation:
\begin{equation}
\sigma_{ij,j} =  f_j  +\frac{\partial\sigma_{j}}{\partial x_{j}} + \sum_{q\neq j}\frac{\partial\sigma_{qj}}{\partial{q}}, 
\end{equation}
with $j=1,2,3$

The terms $\sigma_{ij},\epsilon_{ij}$ represent the stress and strain tensors  and $\kappa_{ijkp}$ is the 6x6 stiffness matrix. $P_p$ is the pressure and $f_i$ is the ith component of the vector of body force per unit volume $\vec{f} = \vec{g}\rho$, where $\vec{g}, \rho$ are the gravity and the mass density.  The eigen-values of  $\sigma_{i,j}$ will be referred as the principal stress components and will be denoted as 
$\sigma_1,\sigma_2,\sigma_3$,  with $\sigma_1 < \sigma_2 < \sigma_3$.

Within the approximation of linear elastic and isotropic solids that will be used here, the 36 components of the stiffness matrix can be described as a combination of two elastic properties: stiffness $E$, and  Poisson's ratio $\nu$, see \citep{timosenko}.

There are several subtleties in solving \ref{eq1} via finite elements that relate with the contents of this paper: multi-scale character of the solutions, non-locality and computational cost.

\subsection{Non-locality and multi-scale character of the stress field}
Non-locality means that the stress calculated at a given point $(x_i ,y_i, z_i)$ depends on the solutions at all the other $(x_j ,y_j, z_j)$ points in the 3D volume studied. As an example, it is known a salt dome disrupts the stress kilometers away from it \citep{koupriantchik}. Hence, the stress field away from the salt body would depend on the local distribution of pressures, and mechanical properties among other variables but it will be also influenced by far-field structures. 

At a smaller scale, stress magnitude varies across intervals of different lithologies along the stratigraphic column. These vertical stress variations (gradients) are due to the presence of relatively thin layers of different lithologies with varying mechanical properties \citep{amadei}. For example, intervals of stiff rocks such as limestone would tend to concentrate more stress than softer rocks, such as shale. Yet lamination and compositional heterogeneity among other variables that act at a much finer scale introduce a high-frequency variability of the stress magnitude inside each lithology. 

\subsection{Computational cost}\label{cost}

Finite element modeling relies on the generation of a grid that subdivides the volume of interest. As the size of the model increases, or as more resolution is needed, the number of cells in the model increases and the computational cost increases linearly (at best). 

To put this in context, consider a 3D model covering a horizontal area of 5 km per side and 10 km in depth with a resolution of 50 m in the horizontal and 1 m along the vertical. The model would require 100 million cells and solving Eq. \ref{eq1} would require a modest computer cluster. The model would cover a relatively small volume of interest with one single horizontal well in the middle. In an exploration scenario the area of study could easily be 30 km per side and the cell count would be above 3 billion elements at the same resolution. At the time of writing this paper, a finite element model of this size would require  a large-scale and high-performance computing infrastructure. To keep the cell count within manageable levels, the grid would need to be refined in specific regions and coarsened everywhere else or some upscaling or downscaling technique would need to be implemented. 

In this paper we propose a method suitable to model stress at different scales and will apply it in an example case where the scale of observation ranges from the sub-meter to the kilometer-scale. Yet the method is not constrained to these ranges. The details of the method will be presented next.

\section{Proposed method}
\subsection{Rationale}
The method proposed relies in a conceptual view of the solutions of the elliptic differential equation in question (in our case the the elasticity equation) in the 3D space as a superposition of at least two contributions. One would be due to the average trends, that act in the large scale. This will be called the low-frequency solution. The second contribution would be a high-frequency component superimposed on the previous one, due to the small-scale heterogeneity of the physical system. 

The low-frequency solution can be obtained via inexpensive finite element simulation at a coarse scale of observation for which the input parameters (rock properties, for instance) would be represented by average trends. Finite element solutions would be compliant with the elastic properties of the model, pressure, tectonics and the equilibrium conditions in Eq. \ref{eq1}. 

Such coarse solution would need a correction to add the high-frequency component of the solution that is present at the fine scale. Knowing how to compute such correction would be equivalent to knowing what transformation would to be applied to the coarse solutions to obtain the solutions at the fine scale.

\begin{figure*}
    \centering
    \includegraphics[width=0.85\linewidth]{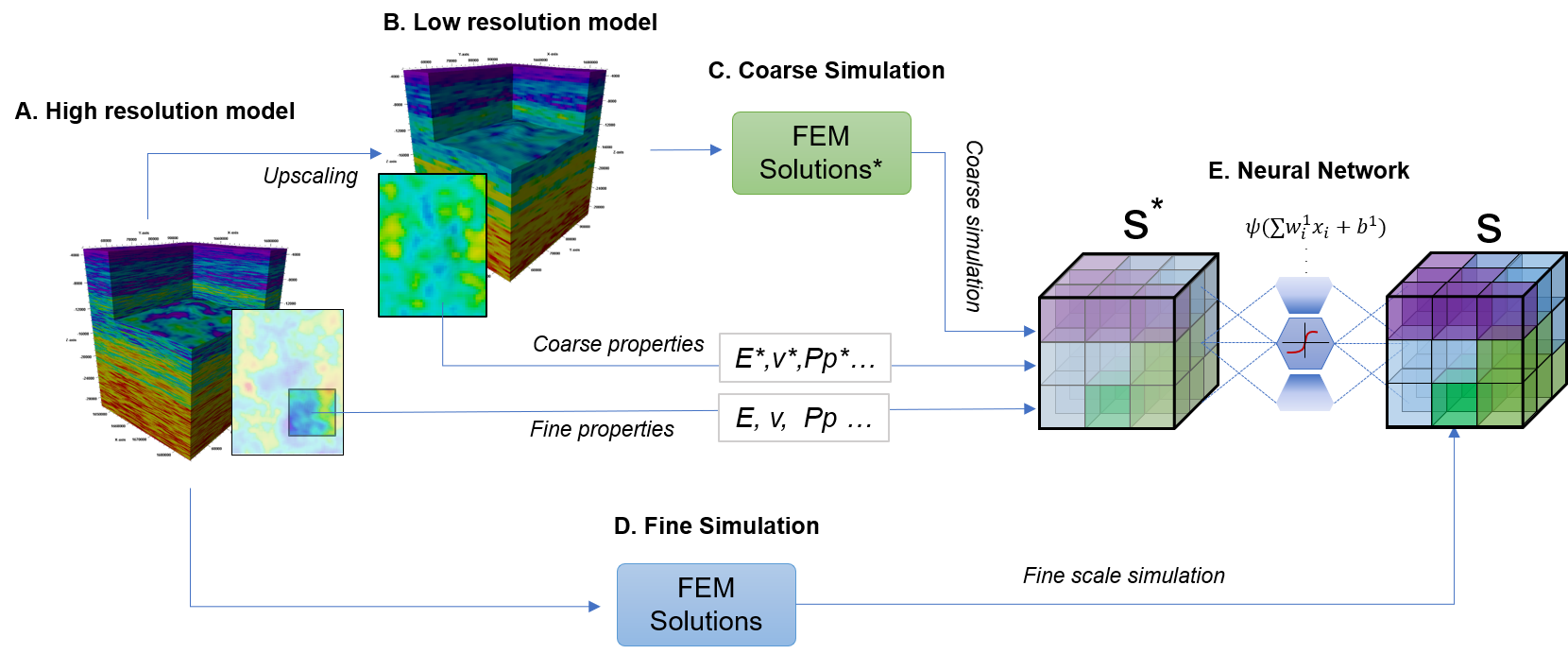}
    \caption{ \scriptsize{Schematic of the method proposed. A: A high resolution model is built including the relevant $x$ features. B: All the features are scaled into a model of coarse geometry. C: Finite element solutions (FEM) are obtained in the coarse model. D: FEM solutions are obtained in a subvolume of the fine scale model. E: Free parameters and solutions at both scales are used to train a neural network to find the relationship between coarse and fine-scale solutions. 
  }  \label{workflow}}
\end{figure*}

A transformation that would perfectly map the finite element solutions in-between different scales, however, may not exist. This is because any transformation would operate on known local information (mech properties both scales, fluid pressure, geometry) to produce a proxy of the stress solution in the fine scale, but the condition of mechanical equilibrium in the fine scale cannot be inferred only based on local variables. 

For instance, consider that the elasticity solution at any given element of the high-resolution model, would need to be in mechanical equilibrium with the adjacent small cells (and all the others). To impose mechanical equilibrium, the partial equation Eq. \ref{eq1} with the appropriate constraints needs to be solved. That’s what the finite element solver does. Altogether, the solutions at the coarse scale $\sigma^*$ would satisfy the mechanical equilibrium constrains, but the transformed solution may not. 

We argue that part of the non-locality information of the fine-scale is still embedded in the coarse solution, which could serve as an approximation. This is because, a given combination of the coarse principal stress components  $\sigma_1^*,\sigma_2^*,\sigma_3^*$ at a specific coarse cell is not arbitrary. That solution reflects the fact that such a cell is in mechanical equilibrium with its neighbors, which in turn are in equilibrium with their neighbors under the specific tectonic forces acting at the global scale and for the global stiffness matrix of the system. Altogether, the coarse stress solutions encode information about the long-rage spatial correlations of the stress field. If this is the case, then we argue that there can be a transformation $f$ that within an acceptable error margin  $\beta$ would map the solutions of the partial differential equations in-between scales:
\begin{equation}\label{eq4}
\sigma_{ij} = f(\sigma^*,x_i )+\beta
\end{equation}

The term $f(\sigma^*,x_i )$ in \ref{eq4} represents such a generic transformation (or function). The terms $\sigma_{ij}$ and $\sigma_{ij}^*$ are respectively the solutions in the fine-scale and the coarse scale. The term $\beta$ in \ref{eq4} is an error, because the transformation between scales is not perfect and $x_i$ represents a set of variables that $f$ operates on.  We propose to use techniques of machine learning to obtain $f$ and the workflow to implement this idea will be discussed next. Please note that in what follows, an asterisk will be used to denote quantities in the coarse scale. For instance, $E$ and $E^*$ will denote the stiffness in the fine and coarse scale respectively.

\subsection{Workflow}
The method starts with a high-resolution model of the subsurface (see Figure \ref{workflow}A), including relevant features of the problem $x_1, x_2 \dots x_n.$ Constructing the model is not a problem in general. The challenge is what to do with that model if solving the elasticity equations becomes prohibitively expensive. This is when a second model comes into play (see Figure \ref{workflow}B). 

The second model is constructed from the high-resolution one but it should be coarse so that  the elasticity solutions can be obtained inexpensively. Yet, it must be fine enough to capture the overall trends. It may become apparent that if the coarse model is excessively coarse, say for example that it is so coarse it has only one cell, it would not capture any relevant information. As shown in Figure \ref{workflow}B, the mechanical properties, pressure, boundary conditions and other relevant features $x_n$ are upscaled (mapped) from the high resolution into a new set of features $x_1^*,x_2^*\dots x_n^*$ at the coarse scale. 

The method proposed here is not attached to any particular upscaling method. However, one would expect that the closer the coarse solution is to the correct effective response, the smaller the amount of information that would need to be encoded in the neural network.

The next step is to obtain a solution $\sigma^*$ to the pertinent equations (Figure \ref{workflow}C). A key step in the workflow would then be how to approximate the fine-scale solution $\sigma$ from the coarse solution $\sigma^*$. Here is when we propose to use a trained neural network. This part of the workflow is sketched in the lower branch of Fig. \ref{workflow}.

The workflow would follow by solving the same set of equations in the fine scale but only within a manageable sub-volume inside the area of interest. This volume must be big enough to capture the relevant heterogeneity, but it must be small enough to allow solving the elasticity equations within pragmatical boundaries of time and computational resources.  Once the solution is at hand (Figure \ref{workflow}E), the neural network would be used to find the transformation between the solutions at both scales, i.e. the term $f$ in Eq. \ref{eq4}. 

Note that this methodology incorporates the particular characteristics of the model studied such as mechanical properties and structural details. Hence, the transformation $f$ will be model-dependent. If found, however, the transformation $f$ could be applied across the 3D volume of interest and recover a solution for stress in the high-resolution space for the specific model studied. The challenge here is to decide what are the relevant features $x_i$, how to obtain such transformation and how to train the neural network. This will be discussed in the forthcoming sections. For the sake of demonstrating the model capabilities, an example physical model that will be used throughout this paper will be introduced first.

\section{Finite element model}
\subsection{Geological context}
 The physical model under the study covers a vertical extent from a depth of 1.2 km below the surface to a depth H=9 km. The model’s horizontal area is approximately 170 x 106 m$^2$. The model includes a fold extending along the North-South direction of the area of interest. 

\begin{figure}[!h]
    \centering
    \includegraphics[width=\columnwidth]{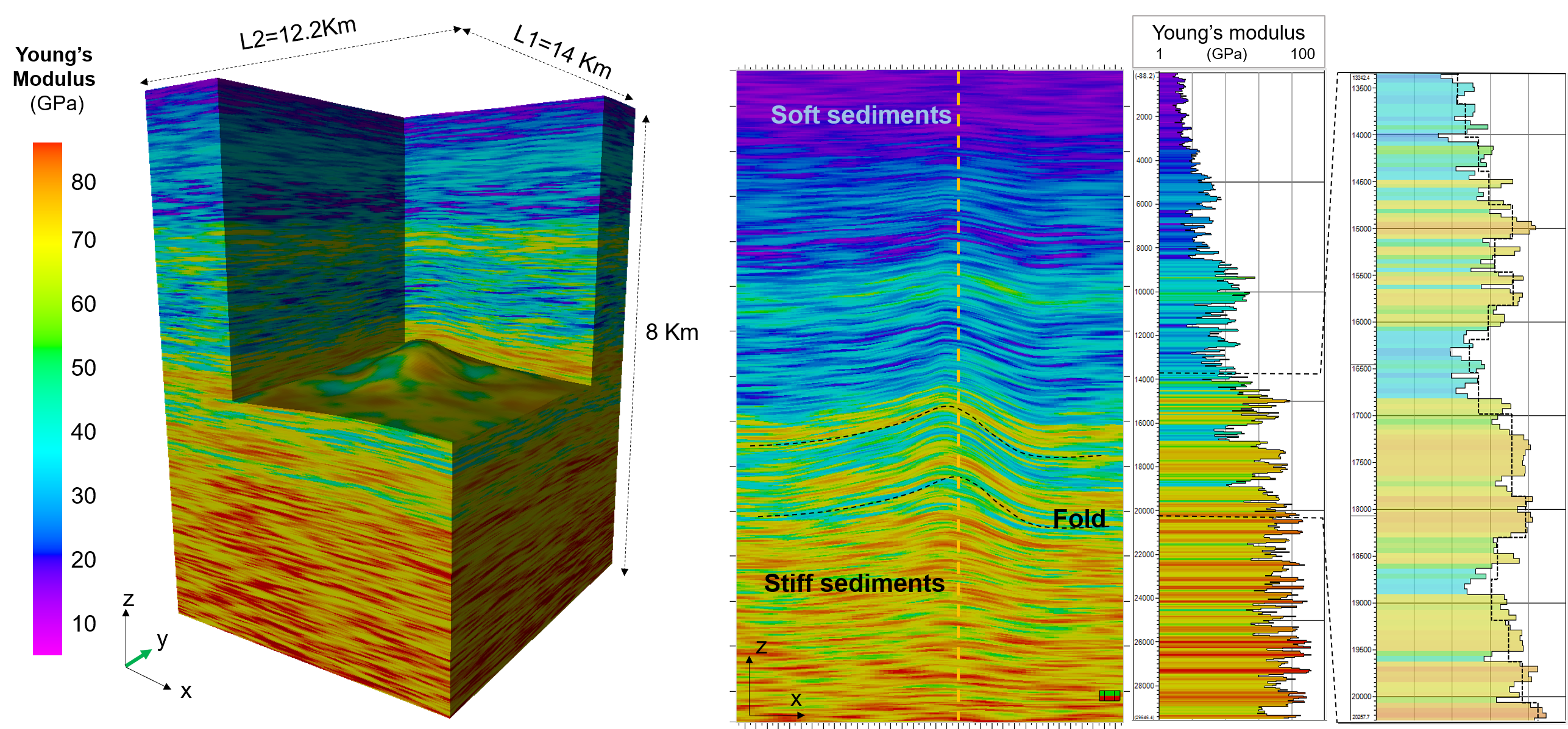}
    \caption{\scriptsize{Example model. 3D color map of rock stiffness (left). Vertical cross section (middle). Stiffness projected along a vertical line (right). The dashed line in the right-most figure is the running average trend.} 
    \label{model}}
\end{figure}

The rock mechanical properties are distributed in the 3D volume within typical ranges of shales and sandstones. These vary along the vertical and horizontal directions across the model. Figure \ref{model} shows the rock stiffness as an example and Table \ref{tab:parameters} indicates the range of variation of several variables across the model. 
\begin{table}[!h]
\centering
\caption{Model parameters}
   \label{tab:parameters}
\begin{tabular}{lcc }
    \hline
    Parameter & Magnitude &Units   \\ 
    \hline
    Young's modulus & 	5 - 85	& GPa\\
    Poisson Ratio	&0.2 - 0.42&	-\\
			Bulk density	&2.1 - 2.5	&gr/cm3\\
			Pressure gradient	&10.63	&kPa/m\\
			Min effective stress gradient	&4.3-7.01	&kPa/m\\
			Tectonic effective stress anisotropy&	1.11 - 1.41&	-\\
			Vertical effective stress gradient	&10.1-12.2&	kPa/m\\
    \hline
\end{tabular}
\end{table}

The model as described is relatively simple when considering the spectrum of scenarios that can be found in geomechanics. One characteristic of it is that it contains no faults. Also, there is a significant variability in the mechanical properties, both vertically and laterally but there is a clear horizontal correlation among these. This persistency of the properties is found in many settings but not in all. Should there be a fault, for instance, this wouldn't be the case. Even with its given benefits in terms of simplicity, porting elastic solutions between numerical models of this system constructed at different scales would be a challenge. This will be shown later in this paper. 

%Note that the stiffness’ depth-trend is non-trivial, e.g. the stiffness shows interbedded intervals of hard or soft rock along the stratigraphic column and within each interval there is a variation around a mean value. 
%It was assumed a Normal stress regime at the regional scale and a hydrostatic pressure gradient. This is a common scenario \cite{fjaer, zoback2007}.
%

\subsection{Discretization and material properties}

The physical model described before was represented in two finite element models differing only in their cell size (resolution). Table \ref{tab2} indicates the geometric details in both cases. 

\begin{table}[!h]
\centering
\caption{Grid geometry in the fine and coarse models}
   \label{tab2}
\begin{tabular*}{\columnwidth}{lcc }
    \hline
Element size (m)	&Fine model&	Coarse model\\ 
    \hline
    Height	&4.5&36.0\\
   Width	&36.6&	73.2\\
    \hline
\end{tabular*}
\end{table}
Note that in the horizontal directions, the edge-length of the elements in the fine model is half of the length of elements in the coarse model. In the vertical direction the element height in the fine model is $1/8$ of the coarse one. With this relation of element sizes, a total of 32 elements of the fine model are fully enclosed inside each element of the coarse model. 

The mechanical properties of the rock were first populated in the fine grid, and then upscaled to the coarse model. The assigned values to each coarse cell were the volume-weighted arithmetic average of the values of each of the fine cells intersecting the coarse cell:
\begin{equation}\label{eq5}
x_i^* = \frac{  \sum_{k=1}^{k=32} x_k w_k } {   \sum_{k=1}^{k=32} w_k    }
\end{equation}

\begin{figure}[!h]
    \centering
    \includegraphics[width=\columnwidth]{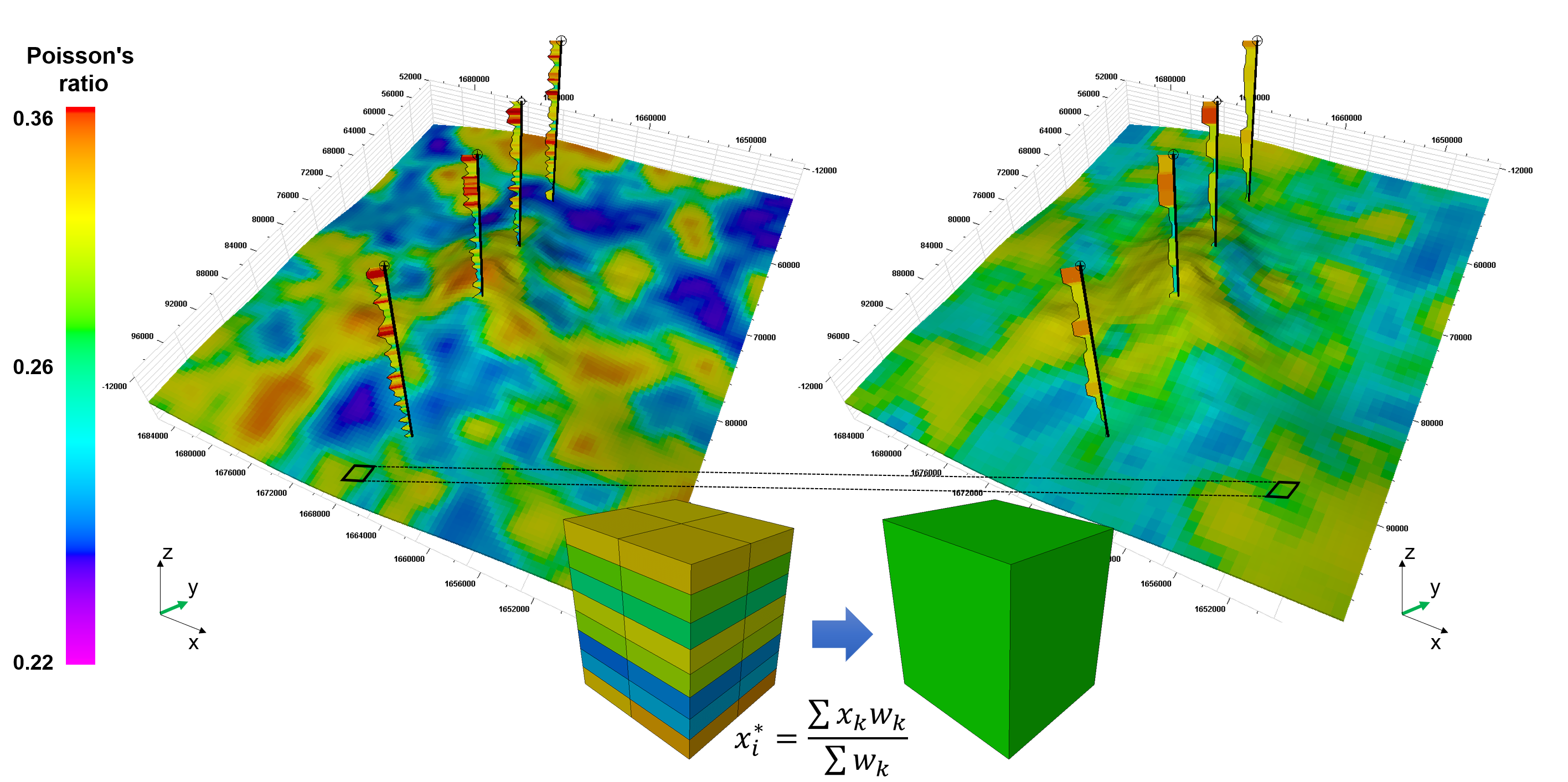}
    \caption{\scriptsize{Poisson's ratio across a 2D slice of 3D model. Fine model (left) and 
    coarse model (right).  The inset at the bottom of the figure shows the cells of the fine model intersected by one coarse cell at different depths and their average.
    }\label{poisson}}
\end{figure}

The term $x_k$  denotes the features (i.e. Young’s modulus, density and Poisson’s ratio) in the k cells of the fine model that intersect the cell ith in the coarse model. The coefficients $w_k$ correspond to the volume of the coarse cell intersected by each of the k fine cells. In the particular case considered here, each element of the fine-scale model is fully contained in one cell of the coarse model. Hence $w_k=1$ in Eq. \ref{eq5}. As an example, Fig. \ref{poisson} shows the representation of the rock Poisson's ratio at the two scales. 

A point to note is that terms such as coarse, fine or high-resolution used in reference to grids are relative. The high-resolution model used here, still has a manageable resolution so that we could obtain finite element solutions in that model. It may be considered coarse for many applications, but it is highly resolved in comparison with the upscaled model that is also used through this paper.

%A point to note is that terms such as coarse, fine or high-resolution used in reference to grids are relative. The high-resolution model used here, still has a manageable resolution so that we could obtain finite element solutions in that model. It may be considered coarse for many applications, but it is highly resolved in comparison with the upscaled model that is also used through this paper.

Equation \ref{eq1} can be solved in the scale of the coarse model to obtain a solution for the stress tensor. The problem is then how to recover the stress in the fine resolution. How, for instance, can we get an estimate of the stress in the thin layers of sand and shale that disappeared during the averaging process? 

In the previous section it was proposed a method that requires to obtain stress solutions via finite element modelling in the coarse model and in a manageable sub-volume of the fine model. This step is described next.

\subsection{Elastic stress solution}
\begin{figure}
    \centering
    \includegraphics[width=\columnwidth]{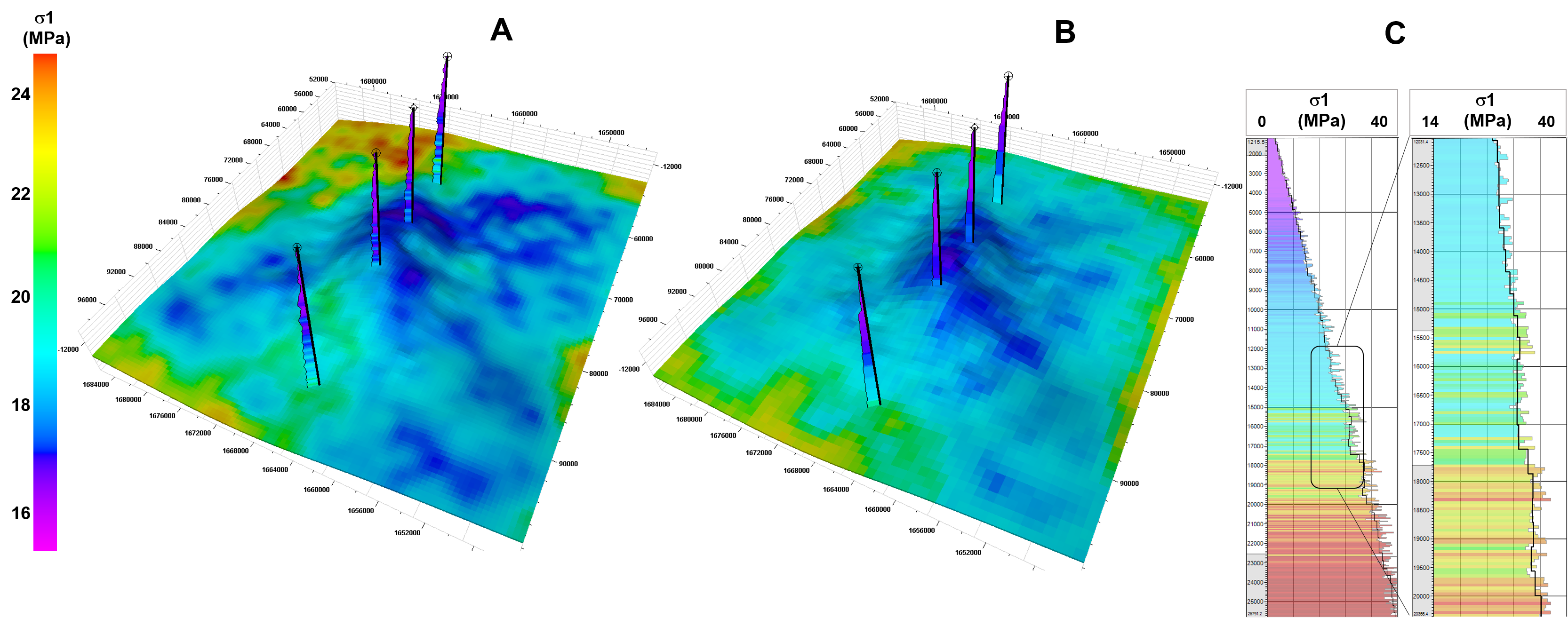}
    \caption{\scriptsize{Effective minimum principal stress $\sigma_1$. A: Cross horizontal section of the fine model B: Cross horizontal section of the coarse model. C: Projection of  $\sigma_1$  along a vertical line and zoomed-in view. Filled curve corresponds to the fine model and the dashed line to the coarse model. 
    }\label{s1}}
\end{figure}

\begin{figure} 
    \centering
    \includegraphics[width=\columnwidth]{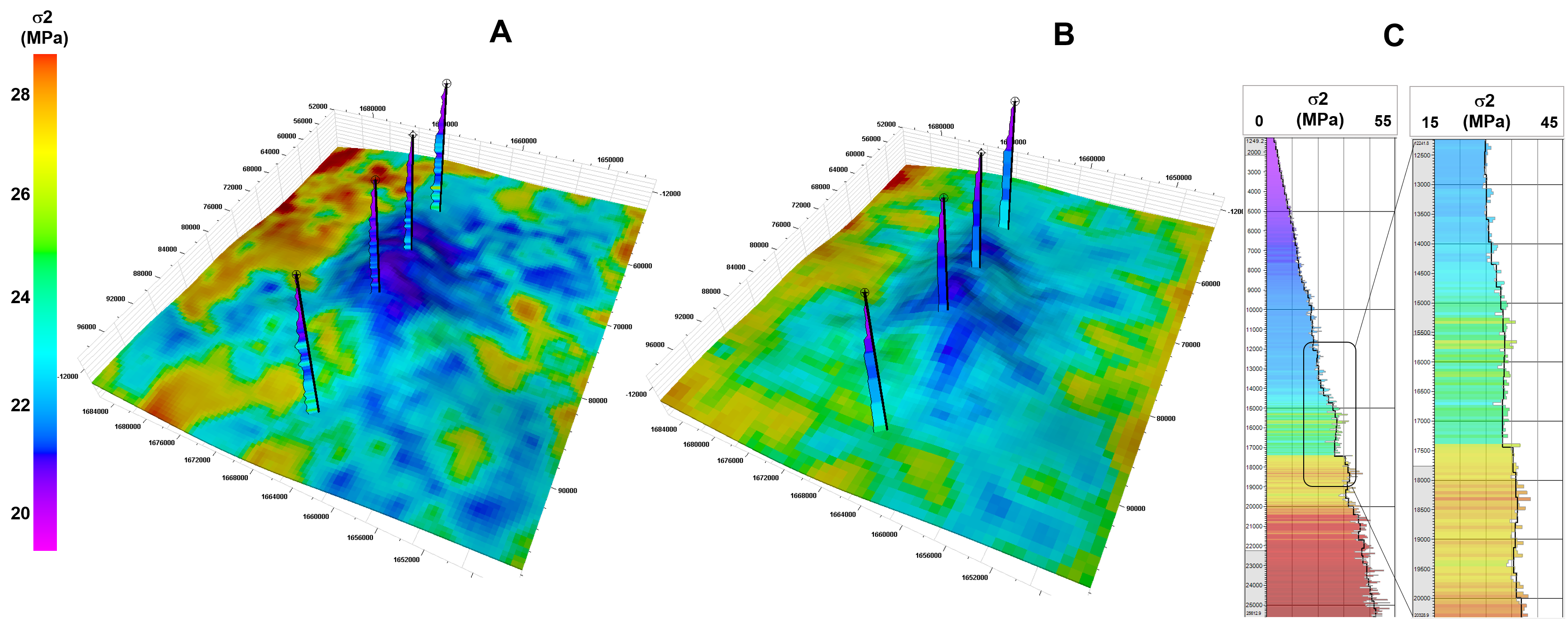}
    \caption{\scriptsize{Cross section of the effective mid-principal stress $\sigma_2$. A: coarse model, B: fine model. The coarse model captures average trends whereas the fine model captures the details. C: $\sigma_2$  projected along a vertical line. Filled curve corresponds to the fine model and dashed line to the coarse one.
    }\label{s2}}
\end{figure}

We first obtain the elasto-static solutions of Eq. \ref{eq1} for the coarse and the fine models under gravitational and tectonic loads. Tectonic forces were modeled with a constant displacement boundary condition applied at the faces of the grids. The magnitude of the equivalent strains applied were of $1.0×10^{-5}$ and $1.5×10^{-4}$ along the East-west, and North-South direction respectively. The base of the model  was fixed along the vertical direction (rollers). 
Figure \ref{s1} and Figure \ref{s2} compare the calculated minimum, $\sigma_1$, and intermediate, $\sigma_2$, principal effective stress components in both models. There it is shown that the stress solution at the coarse scale captures the overall trends but cannot resolve the stress heterogeneity captured in the fine model.
%
%For instance, in Fig. \ref{s1}  in both models there is a relatively low stress concentration on top of the southern part of the fold and a relatively high concentration of stress towards the East of the 3D volume. Yet, the fine-scale model shows noticeably more heterogeneity in the horizontal direction. The vertical trends are shown in Figure \ref{s1}C for both models. As before, the fine model captures a high-frequency variability in the solution that is not captured in the coarse model. Such variability arises from the heterogeneity in the rock properties: stiffness E and Poisson’s ratio $\nu$.

The solutions at the coarse scale will be considered part of the free parameters of the neural network model. These are sometimes called predictors. The solutions at the fine scale will be the dependent variable, also called the target variable or predictand.

\section{Neural network model}
\subsection{Training set}

\begin{figure}
    \centering
    \includegraphics[width=0.8\columnwidth]{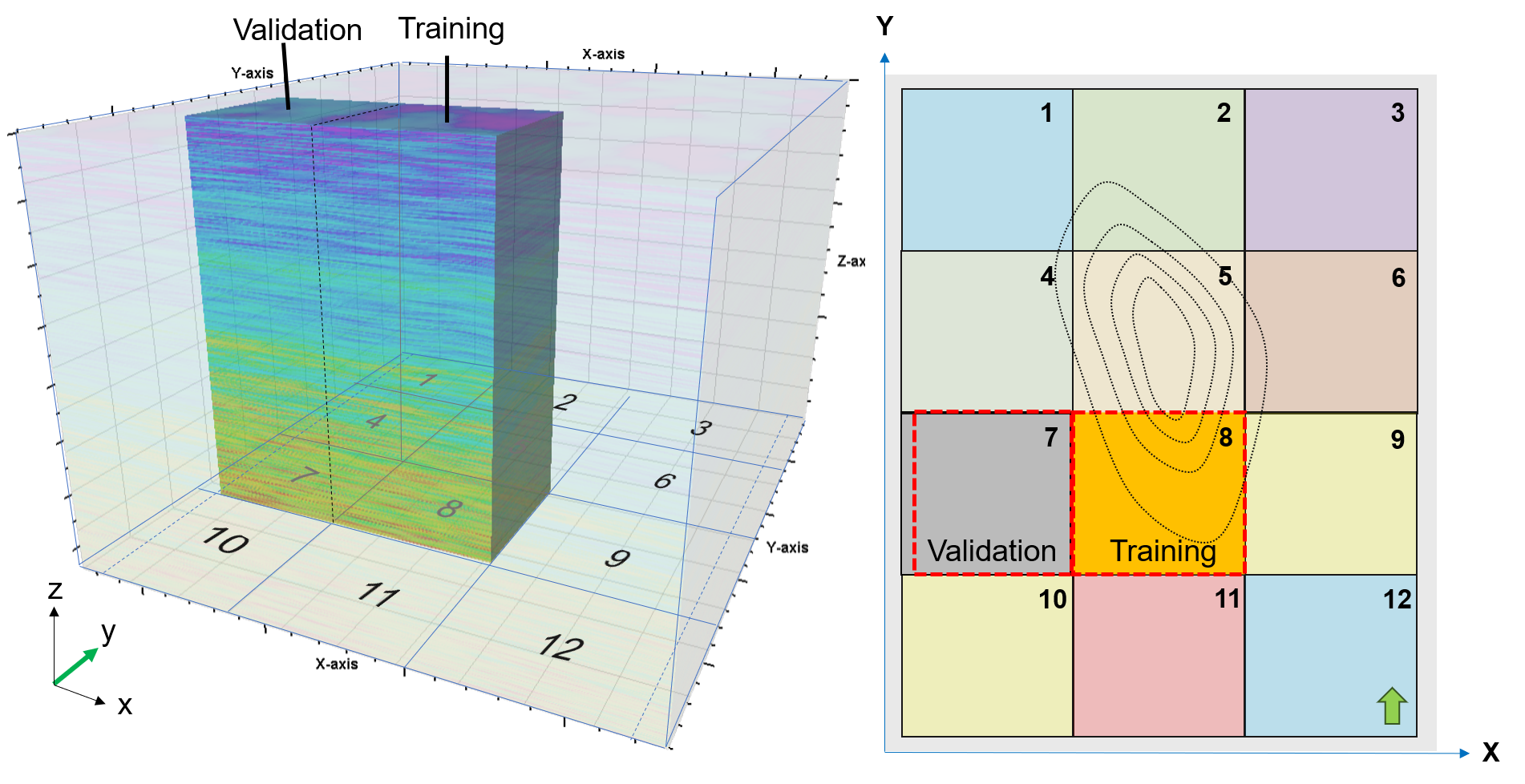}
    \caption{\scriptsize{The 3D volume was divided into 12 squared columns of the same size, extending from top to bottom of the model. Column number 8 was selected for training the neural network. The concentric rings indicate the edges of the fold. }
    \label{trainmodel}}
\end{figure}

The entire 3D model was divided in 12 columns of the same size (see Fig. \ref{trainmodel}) with the intention of training the model with the data of one (or more) columns and using samples from a nearby column for validation. The height of the columns covered the full vertical extent of the model, except for two small gaps at the top and bottom of the model that were discarded to avoid potential boundary artefacts coming from the solutions of the finite element models.

To account for the lateral variability, the correlation length of the mechanical properties in the horizontal plane was estimated via the spatial autocorrelation function of the rock stiffness as $\approx$ 1.5 km \citep{maraunwidmann2018}. The transversal area of the columns was then selected so its horizontal edges were twice as long. With this sub-division, each column enclosed over 470K cells. Results from numerical simulations to be shown later indicate that less than half of this number of samples would be enough to train the neural network used in this work. 

Column 7 was selected for validation. Column 8 was selected for training because it intersects part of the fold, it is not close to the boundaries of the model (see Fig. \ref{trainmodel}) and it is kilometers away from other areas that we can use for comparison later. Although column 5 could have been selected instead, the choice was made for column 8 since selecting column 5 for testing, and not training, appeared as a harder test for the method. One reason is that column number 5 intersects fully the area of greatest curvature of the fold, which is where one would anticipate greater changes in stress intensity and orientation. Leaving this outside the training explores the capabilities of the network to resolve these features without “seeing” them during training.

\subsection{Metrics}
 The terms $\hat{\sigma}_i$ will denote the target value of the ith component of the stress tensor and $\sigma_i$ will denote the prediction. Their difference will be denoted as: $\Delta \sigma_{i} = \sigma_i - \hat{\sigma_i}$ and $\Delta \sigma ^2 = \sum_i{\Delta \sigma_i ^2 }$. In all the numerical experiments done in this work the objective function minimized during training was the mean squared error $ mse = \langle \Delta\sigma ^2 \rangle$. 
The metric used when comparing models in this work was the root mean squared error, $rmse.$ However, other metrics such as simply $\Delta\sigma_i$ in plain units of pressure, or the mean absolute percent error $mape$ will be also used when presenting the results as we considered these more intuitive.

\subsection{Free and dependent parameters}
The main objective of this work is to find the transformation $f$ in Eq. \ref{eq4}. Our hypothesis is that the physical variables involved in such transformation include, the mechanical properties at both scales, pressures at both scales and the coarse-grained stress $\sigma^*$:

\begin{equation}
\sigma(x,y,z)=f(\sigma^*,E,E^*,\nu,\nu^*,P,P^* ). 				
\end{equation}

The term $\sigma(x,y,z)$ in the equation above will refer in what follows to as the tectonic principal stress components, $\sigma_1,\sigma_2$. The maximum principal stress $\sigma_3$ was dropped from the target variables because it is trivial in the case considered here. It does not depend on tectonics or mechanical properties and it is similar in the coarse and fine models. The focus will be on the tectonic stresses, $\sigma_1,\sigma_2$. The min principal component, $\sigma_1$, generally called the fracture gradient is arguably the most important principal component of the stress tensor in geomechanics. 

Some variables that appear in Eq. \ref{eq1} were discarded as predictors because they are redundant. Hook's law in Eq. \ref{eq1} relates strain and stress. Hence, we argue that one of the two tensors, $\sigma^*$ or $\epsilon^*$ can be dropped. We kept $\sigma^*$  since the target variable is the stress, $\sigma$. 

One would expect that the depth-integral of the rock mass density would contribute to the magnitude of all the principal stress components. Yet we argue that such contribution is already encoded in $\sigma_3^*$  so $\rho, \rho^*$ were dropped. 

\begin{figure}
    \centering
    \includegraphics[width=\columnwidth]{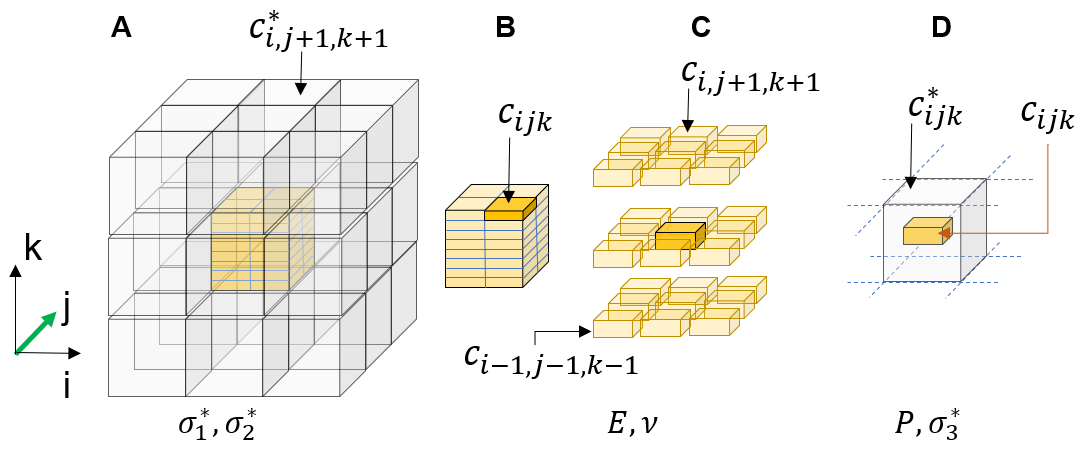}
    \caption{\scriptsize{Enclosing cell and neighbor cells. A: 27 cells of the coarse finite element model. One coarse cell is highlighted. B: Single coarse cell enclosing 32 fine-scale cells. C: Fine cell $c_{ijk}$ and 27 neighbors in the fine-scale model. D: Fine cell $c_{ijk}$ inside its  enclosing cell $c_{ijk}^*$. }
    \label{cells}}
\end{figure}

The mechanical properties were combined into their differences, $\Delta E= E - E^*$, $\Delta\nu=\nu-\nu^*$  because in our understanding, the difference between the stress solutions at different scales is related to these differences in the mechanical properties and not necessarily to their absolute values.
For instance,  each training example will refer to one cell $c_{i,j,k}$ extracted from the fine model and considered as part of the training set. For that cell we will consider the 27 neighboring small cells $c_{i\prime,j\prime,k\prime}$ 
as in Fig. \ref{cells}C. For each of these neighbors, there will be one predictor related to stiffness computed 
as 
$\Delta E = E-E^*$ where $E$ is sampled form the cell 
$c_{i\prime,j\prime,k\prime}$ and 
$E^*$ is sampled from the coarse cell 
$c^*_{i\prime,j\prime,k\prime}$ that contains $c_{i\prime,j\prime,k\prime}.$ Same procedure is followed for $\Delta\nu.$
 
\begin{figure*}
    \centering
    \includegraphics[width=0.88\linewidth]{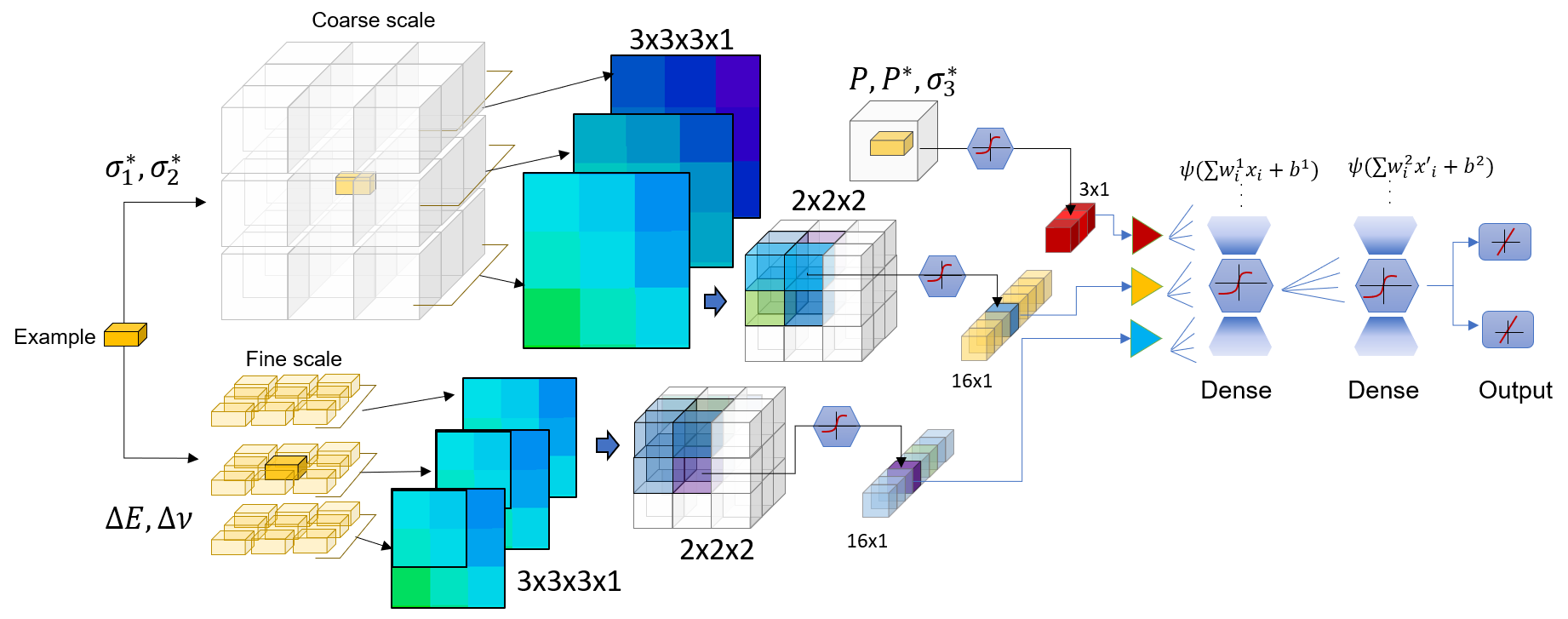}
    \caption{\scriptsize{Network architecture. Four independent 3D valid convolutions using a (2,2) kernel operated on $\sigma_1^*,\sigma_2^*,\Delta E,\Delta \nu*$. The outputs of the convolutions were flattened and merged with $P^*,\sigma_3^*$ and passed to a stage of fully connected layers (see text). }
    \label{architecture}}
\end{figure*}

A fundamental variable is the coarse-scale stress $\sigma^*$ because it encodes the non-local information and correlations between stress, strain, tectonics and mechanical properties in the volume of interest. To account for this information in the training, each training example included the coarse stress $\sigma_1^*,\sigma_2^*$  sampled from 27 neighboring coarse cells $c_{i\pm1,j\pm1,k\pm1}^*$ around each cell $c_{ijk}$  of the fine-resolution model that was included in the training set (Fig. \ref{cells}A). 

%The local variability of the mechanical properties was included by sampling the stiffness $E$ and Poisson’s ratio $\nu$ in the neighborhood of 27 fine-scale cells around $c_{ijk}$ (Fig. \ref{cells}C). 
Pressure $P^*$ and the overburden stress $\sigma_3^*$ show little variability in our model so these were only sampled from the coarse cell $c_{ijk}^*$ containing the fine-scale cell $c_{ijk}$ (Fig. \ref{cells}D).

\subsection{Network architecture}
Figure \ref{architecture} sketches the neural network architecture used in this work.
The network included a convolutional first stage, followed by two densely connected hidden layers and an output layer. In the first stage, the components $\sigma_1^*,\sigma_2^*$ of the coarse-scale stress and the mechanical properties $\Delta E,\Delta \nu$ were mapped into four 3×3×3 matrices and each was passed through an independent valid 3D convolution using a single filter and a 2×2×2 kernel. The output of each convolution was flattened and merged  with $P,P^*,\sigma_3^*$ into a single 1D array of length $N_f=35$. The later was the input to the dense layers in the next stage of the network. The number of neurons in each of those dense layers was  $N_f+5$ and the activation function was the hyperbolic tangent except for the output layer, which was linear. 

Note that one of the advantages of adding the convolutional stage is that it allows sampling the predictors from a neighborhood around the location of each target cell at a relatively low cost. Only eight parameters plus a bias term are involved in the convolution of the 27 stress predictors for each component of the stress. Pooling was not used here but the combination of pooling and different kernel sizes allows to control the number of entries that reach the densely connected layers in the last stage of the network. This can become more important if more neighboring cells were to be considered. For more details on pooling and convolutional layers, see \citep{Goodfellow-et-al-2016, Chollet2017}.

\section{Results}
\begin{figure}
    \centering
    \includegraphics[width=\columnwidth]{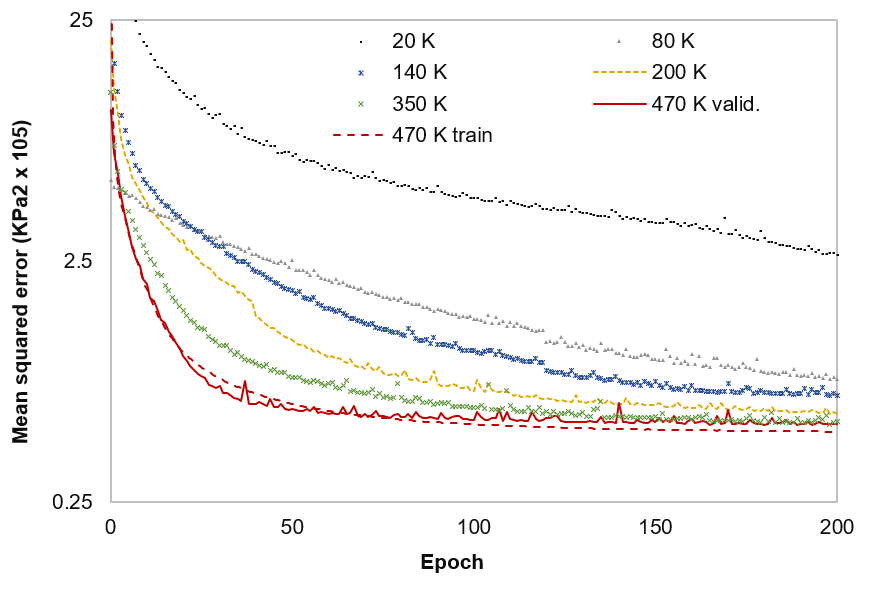}
    \caption{\scriptsize{Evolution of the mse (loss) during training. Curves correspond to the loss in the validation set for different sizes of the training set. For the case of 470K samples, both the training and validation loss are shown.  }
    \label{convergency}}
\end{figure}

The neural network was trained with stochastic gradient utilizing the Keras framework 
available in \cite{Chollet2015}. TensorFlow was used at the backend \citep{tensorflow2015-whitepaper}.
See also \cite{gulli2017deep} and \cite{atienza} for more details. 
Figure \ref{convergency} shows that convergence was achieved after $\approx$100 epochs when trained with the full set of 470K samples (and a batch size of 32). Similar trends were obtained for varying number of examples taken randomly from the 470K available inside the sub-volume selected for training.

The  results show that as the size of the training set increased, the mean squared error computed for the validation set decreased. Yet when training the network for 200 epochs or more, the gains were marginal after including more than $\approx$200K randomly selected samples from the training set. 

Based on this observation, it was concluded that our initial guess on the size of the training set was adequate. The results reported in what follows correspond to the model trained for 120 epochs with all the 470K samples inside sub-volume 8 highlighted in Fig. \ref{trainmodel}.

 %\begin{figure}[b]
    %\centering
%    \includegraphics[width=\columnwidth]{convergency1.png}
%    \caption{\scriptsize{Evolution of the mse (loss) during training. Curves correspond to the loss in the validation set for different sizes of the training set. For the case of 470K samples, both the training and validation loss are shown.  }
%    \label{convergency}}
%\end{figure}

\begin{figure}%\vspace{-0.5cm}
    \centering
    \includegraphics[width=\columnwidth]{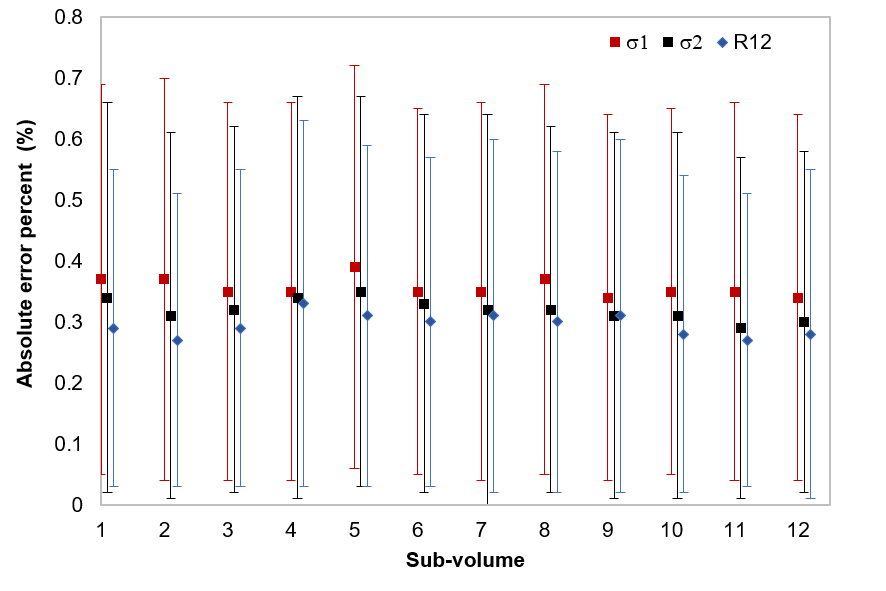}
    \caption{\scriptsize{Absolute percent error across the model. Symbols represent the mean values for each sub-volume. The length of the error bars corresponds to twice the standard deviation. }
    \label{rmse}}
\end{figure}

Figure \ref{rmse} shows the mean absolute error percent calculated for each column. Not a single sector in the model stands out the rest in the figure. In every sub-volume, the mean absolute percent error was within a 0.4\% for either of the two tectonic components of the stress, $\sigma_1,\sigma_2$. However, the error in the estimation of the fracture gradient, $\sigma_1$, was slightly higher than for $\sigma_2$. The error percent in the estimated stress ratio $R_{12}= \sigma_2/\sigma_1$ resulted relatively low in comparison. Our interpretation is that this is because  $\sigma_1$  and $\sigma_2$ are correlated. 

To put these results into context, consider that in absence of depletion, fractures, faults or major structural details, the vertical stress would be proportional to the integral of the density of the rock, $\rho$. Density can be measured via logging tools within a measurement error in the order of 0.05 gr/cm3 to 0.1 gr/cm3 \citep{SmithOnline}. For an average density of the rocks in crust of 2.5 gr/cm3, then the error in the density logging would be in the order of 2\% to 4\%. Hence, an analytical calculation of the vertical stress could be off by at least $\pm 2\%$ in that idealized scenario.

The tectonic components cannot be calculated in such a simple way. These depend on the mechanical properties, and tectonic forces. The minimum principal stress can be estimated in the field within an error of the order 10\% in a favorable scenario \citep{amadei}. Note that scattered measurements of $\sigma_1$  are often used to calibrate geomechanical models. 

Altogether, it appears that in the very best scenario, an error of the order of 2\%-4\% is indeed small for the problem in question in a practical scenario. An average estimate of the order of 0.4\% would represent a minor uncertainty in addition to a standard scenario where a high-resolution model is used to estimate the stress field. 

%%%%%%%%%%%%%%%%%%%%%%%%%%%%%%%%%%%%%%%%%%%%%%%%%%%%%%%
%%%%%%%%%%                            HORIZONTAL                                   %%%%%%%%%%%%%%%%%%

\begin{figure*}%\vspace{-0.5cm}
    \centering
    \includegraphics[width=0.95\linewidth]{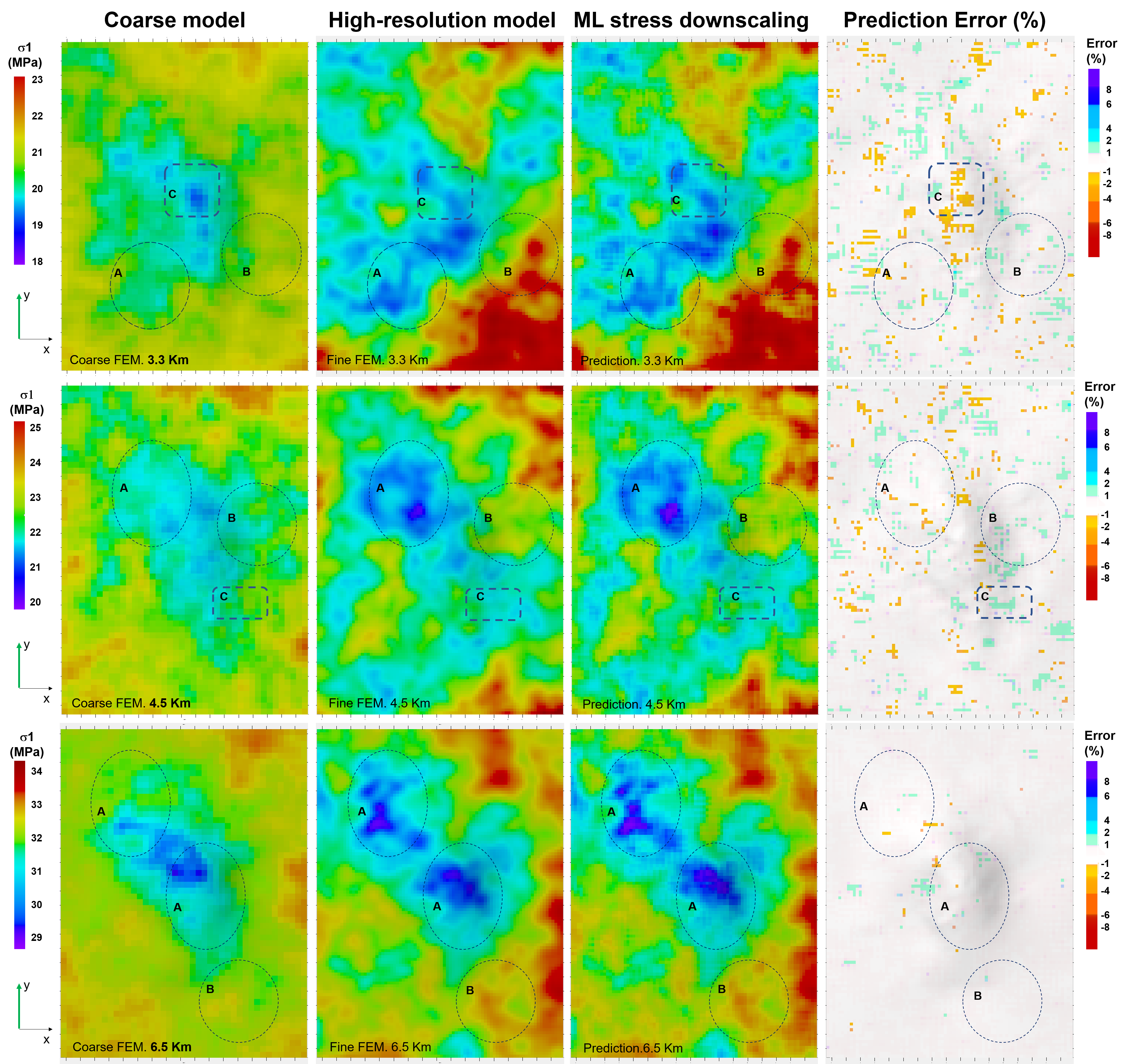}
    \caption{\scriptsize{Fracture gradient calculated in the upscaled model (left column), high-resolution model (middle column) and stress prediction (right column). The three rows from top to bottom show the results at depths of 1/3 ,1/2 and 3/4 of the model depth. The upscaled model over-estimates the fracture gradient in areas labeled as A, and under-estimates it in the areas labeled as B. Even though that model was used as input during the training, the ML downscaling method reproduces the details of the high-resolution solution. Dashed squared are added as a guide.  
    }
    \label{s1_horizontal_view}}
\end{figure*}

Figure \ref{s1_horizontal_view} compares the fracture gradient in the coarse model, the high-resolution model and the stress predictions using the method proposed in this paper. The figure corresponds to a projection of the fracture gradient on horizontal planes at the depths of 
3.35 km, 4.57 km and 5.8 km. These correspond to $\approx$ 1/3, 1/2 and 3/4 of the model's total depth respectively. The entire 
model's horizontal cross-section is shown in the figure.

The results in Fig. \ref{s1_horizontal_view} show that the high-resolution model captures the stress heterogeneity along the horizontal direction. Such heterogeneity is also partially captured in the upscaled model but the model lacks of the needed resolution to resolve important details. For example, in the areas  labeled A in Fig. \ref{s1_horizontal_view}, the fracture gradient is relatively low. In an unconventional reservoir, these could be candidates for hydraulic stimulation. Yet if wrongly flagged as having a relatively high fracture gradient, they could be discarded when searching for sweet spots. 

Note that the solution obtained with the method proposed here approximates well the results obtained by the high-resolution model, which is taken as the correct solution. In the top set of images in Fig. \ref{s1_horizontal_view}, for example, the correct solution is be estimated within an error of less than 2\%. The results in Fig. \ref{s1_horizontal_view} show that not only the magnitude of the stress field is well represented in the downscaled model but also the shape of the regions of high/low stress concentration. If the solutions were available only in the upscaled model, due to computational cost for instance, the fracture gradient could be over-estimated in the regions enclosed in the ellipsoids labeled A and under-estimated in those labeled as B.

Figure \ref{s1_horizontal_view} also shows the (signed) percent of error in the estimate of the fracture gradient. From the images it is apparent that the cells where $\sigma_1$ is estimated with a percent of error of magnitude greater than 1\% are distributed in scattered patches across the model. Visual inspection of the image indicates that $\sigma_1$ can be either over or under-estimated at different locations but a bias in either direction is not obvious. 

The results shown in Fig. \ref{s1_horizontal_view} also suggest that the percentwise accuracy of the stress predictions improves with depth. This is because the average values of the fracture gradient increase faster with depth than the magnitude of the differences between the predictions and the target variable $\Delta \sigma_1 = \sigma_1-\hat{\sigma}_1$. Hence, the percent of error 
$100\Delta \sigma_1/\sigma_1$ decreases. In the particular case of the model considered here, $\sigma_1$ increases from nearly zero at surface to about 45 MPa at the depth of 9 km. Meanwhile the peak-to-peak range of variation of $\Delta \sigma_1$ hardly reaches 130 kPa in the entire model. 

Altogether the results compiled in this experiment indicate that departing from a coarse solution obtained via finite element modeling, and a partial solution obtained in a model with high resolution, the technique presented can reconstruct the fracture gradient field in 3D at the resolution of the finer model. The results in Fig. \ref{s1_horizontal_view}  indicate that percentwise differences between the predicted fracture gradient and the correct result are within the range of 2\%. 
%%%%%%%%%%%%%%%%%%%%%%%%%%%%%%%%%%%%%%%%%%%%%%%%%%%%%%%%%%
%%%%%%%%%%%%%%%%%%%%%%%%%%%%%%%%%%%%%%%%%%%%%%%%%%%%%%%%%%

%%%%%%%%%%%%%%%%%%%%%%%%%%%%%%%%%%%%%%%%%%%%%%%%%%%%%%%%%%
%%%%%%%%%%%%%%             VERTICAL                                        %%%%%%%%%%%%%%%%%%%%%%
\begin{figure*}[t!]%\vspace{-0.5cm}
    \centering
    \includegraphics[width=0.95\linewidth]{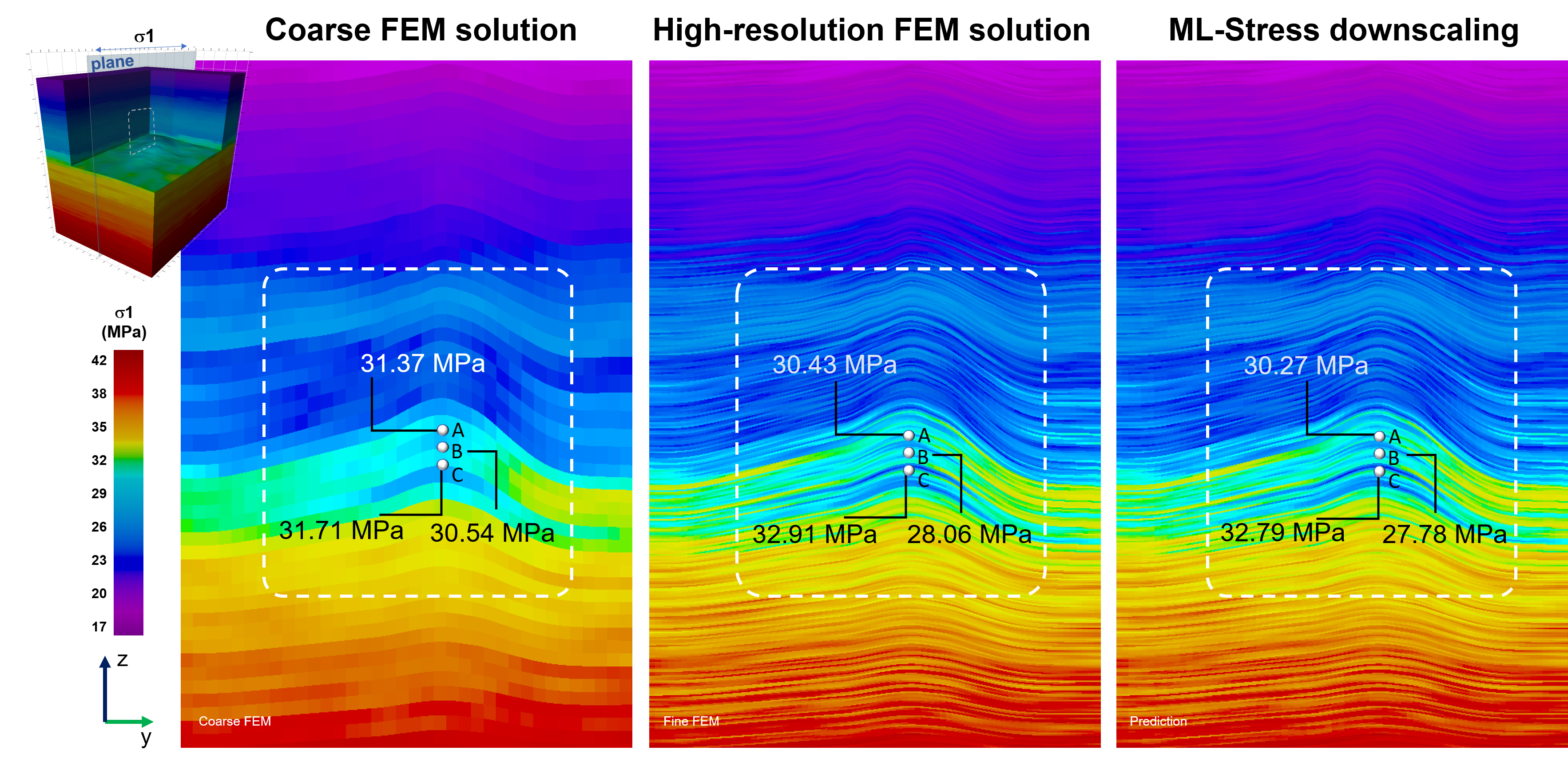}
    \caption{\scriptsize{Zoomed-in view of the fracture gradient projected on a vertical plane. }
   \label{verticalview}}
\end{figure*}

The results shown in Figure \ref{verticalview} correspond to a projection of the fracture gradient on a vertical plane oriented in the direction of the minimum compressive tectonic stress, $\sigma_1$. The image is a close-up showing approximately 1/3 of the total plane height. 
The results indicate that the high-resolution model captures a significant stress heterogeneity along the vertical direction. The contrast in the magnitude of $\sigma_1$ between adjacent layers cannot be well represented at the resolution of the upscaled model. Even though this was the upscaled model used in the training of the neural network, the stress prediction in the latter appears to closely approximate the fine detail observed in the high-resolution model.

For the cells highlighted in Figure \ref{verticalview} as an example, the difference $\Delta \sigma_1$ between the magnitude of the fracture gradient estimated via the neural network and the value computed in the high-resolution model is within $\approx$0.28 MPa. Note that at the depth shown in the figure, the magnitude of the fracture gradient is in the order of 28 to 33 MPa. Percentwise, 0.28 MPa would represent an error of $\approx$1\%. For the two cells, labeled A and B in Fig. \ref{verticalview}, the high-resolution model estimated a stress contrast of about 4.85 MPa. This stress contrast is not resolved in the upscaled model. Yet, the estimated contrast  was 5.01 MPa with the method proposed here, i.e. a 3\% difference. 

These results highlight the potential of the method proposed here: From the finite element solutions obtained in a coarse model such as the one shown in Fig. \ref{verticalview}, and a partial solution of the stress obtained in a high resolution as in the middle frame of  Fig. \ref{verticalview}, the stress field is reconstructed in the 3D space at high resolution as in the right frame of Fig. \ref{verticalview}.

As the estimation of the minimum principal effective stress $\sigma_1$ is of practical importance, it is also  important to estimate the relation between  the magnitudes of the different stress components.  As an example, it is important in the assessment of the potential failure of deviated wells \citep{fjaer,zoback2007}.  

\begin{figure}[b]
    \centering
    \includegraphics[width=\columnwidth]{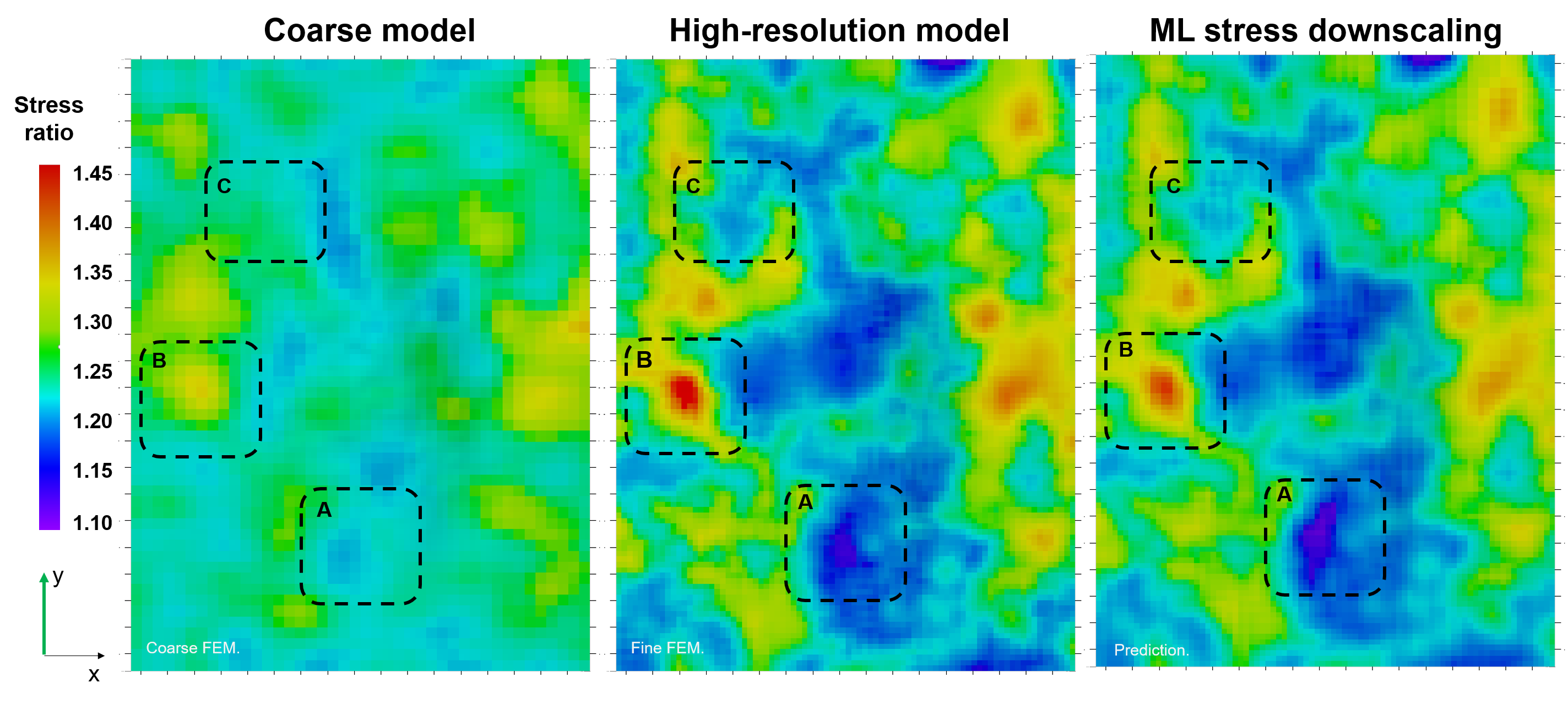}
    \caption{\scriptsize{Stress ratio $R_{12} = \sigma_1/\sigma_2$ for the coarse model (left), the high-resolution model (middle) and result of the downscaling process (right). Dashed boxes as added as a guide (see text).}
    \label{stressratio1}}
\end{figure}

Figure \ref{stressratio1} shows the results obtained in relation to estimated ratio of the principal effective stress components $R_{12}=\sigma_2/\sigma_1$.  The image shows $R_{12}$ projected on the same horizontal plane taken as example in the top row of Fig. \ref{s1_horizontal_view}.

Visual inspection of the results in Fig. \ref{stressratio1} indicates that $R_{12}$ estimated via the ML-stress prediction  is close to the high-resolution finite element results. For instance, the areas with a relatively low or high stress ratio labeled as A or B in the figure, seem to be well represented even when little contrast in $R_{12}$ is captured inside those areas in the upscaled model used in the training. The box labeled C in the figure highlights obvious differences between the correct stress ratio and the estimated one. These differences, however, are relatively small. Aside from a few scattered cells, the cell-by-cell difference between the stress ratio $R_{12}$ estimated in this work and the correct result 
is within the range of $\pm 0.02$. This is 55 times smaller in magnitude than the smallest stress ratio of 1.10 in the entire 3D model. This would correspond to a percentwise error in the order of $\pm$ 2\%. Clearly, the model after downscaling represents a clear improvement over the initial upscaled model used for training.

\begin{figure}[h!]
    \centering
    \includegraphics[width=\columnwidth]{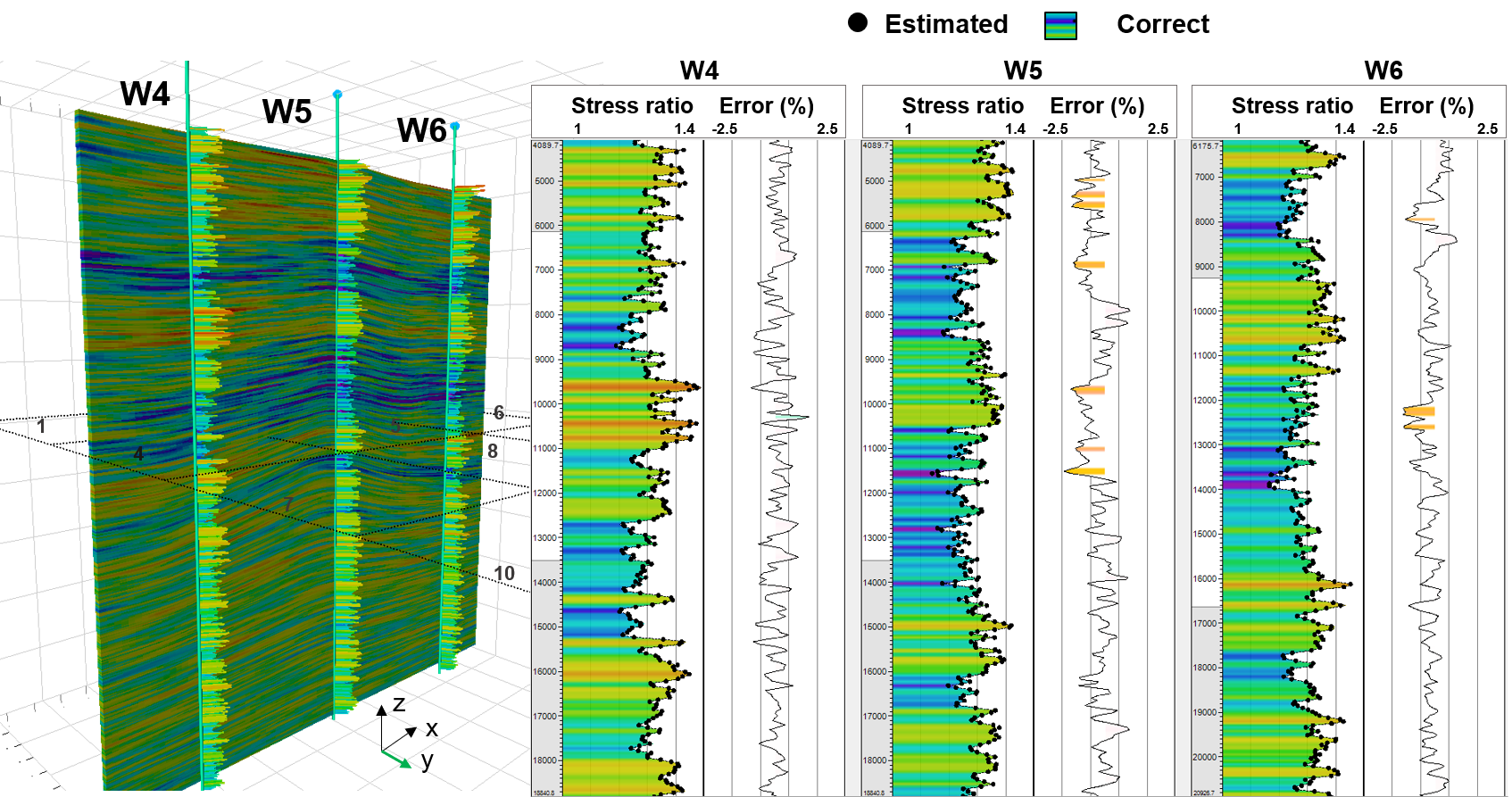}
    \caption{\scriptsize{Vertical projection of the stress ratio along three hypothetical wells centered in sub-volumes 4,5 and 6. Filled circles correspond to the estimated stress ratio and the filled curves to the correct result. 
    }\label{wells}}
\end{figure}

Finally, Fig. \ref{wells} compares the vertical projection of the estimated stress ratio along 3 hypothetical wells centered at sub-volumes 4,5 and 6 as defined in Fig. \ref{trainmodel}. Note from the synthetic logs in the figure that the approximation done by the method proposed here differs from the high-resolution model is about 2\% at most for all the three wells. We argue that this would be a reasonable approximation for most practical purposes.

 \section{Comparison with alternative scaling techniques}
 The method proposed in this work follows the structure of a family of stress downscaling techniques proposed by \cite{herrera}. In that reference the authors compute elasto-static finite element solutions at a coarse resolution and then downscale the stress solutions provided that the fine-scale mechanical properties are consistently used in the coarse model. 

A simple instance of those techniques would be, for example, to estimate the strain $\epsilon_{ij}^*$ at the coarse scale using finite element methods and then insert it into Eq. \ref{eq1} to estimate stress analytically, but this time using the fine-scale stiffness matrix. This would correspond to applying Hook's law at the fine scale but with strains approximated via simulation at a coarser scale. Note that unless explicitly obtained via simulation, the strain tensor would be unknown. In what follows we will refer to the method proposed in \cite{herrera}  as constant-strain downscaling.

The constant strain downscaling also attempts to exploit the fact that coarse strains encode the large-scale and non-locality of the solutions. The difference with the approach taken in this paper is that when the coarse strain is inserted in Eq. \ref{eq1}, the transformation function does not attempt to preserve static equilibrium between the fine stress results. In this paper the transformation is found using techniques of machine learning in the hope that we can learn the relationships between equilibrated stress solutions and the underlying mechanical heterogeneity of a medium.  

\begin{figure}[b]
    \centering
    \includegraphics[width=\columnwidth]{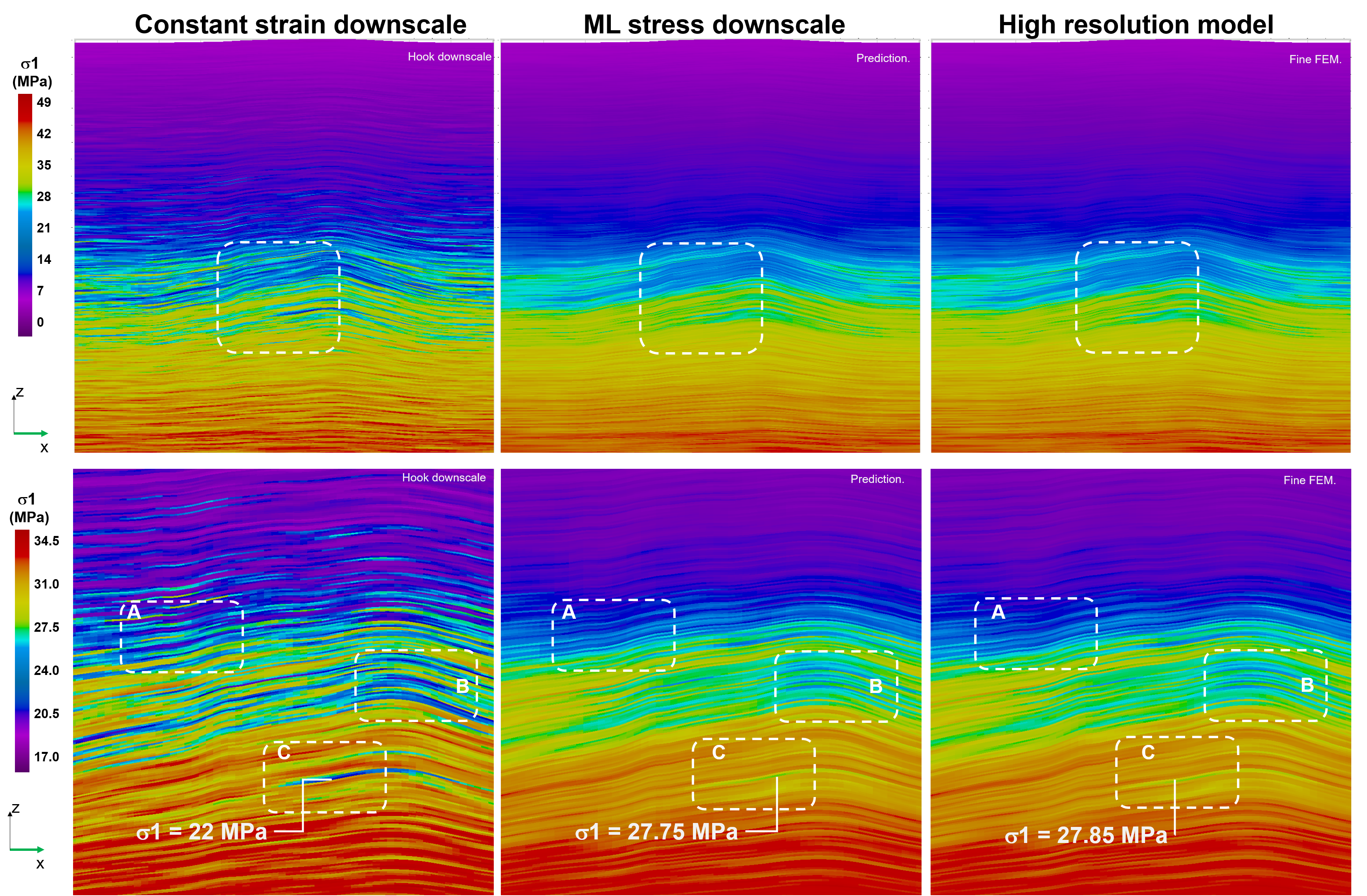}
    \caption{\scriptsize{Comparison between the constant strain downscale method \citep{herrera} and the ML stress prediction proposed in this paper.
    }\label{straindowscale}}
\end{figure}

Figure \ref{straindowscale} compares the minimum principal stress obtained by the constant-strain method and the method proposed here. The figure shows a projection of the minimum principal stress on the plane parallel to it and passing through the center of the model.
The results show that the constant-strain downscale method captures a significant amount of detail of the structure of the stress field. Note for instance the layering in box A and the relatively low stress area around the local bending in box B. The one observation in the figure is that in the case of the constant-strain downscale method, the variability of the stress field appears exaggerated. The figure highlights as an example of one layer in box C where the magnitude of $\sigma_1$ is under-estimated by nearly $\Delta\sigma_1\approx$ 5.8 MPa. This is an error in the order of 20\%, i.e., about 1 order of magnitude higher than the average error obtained with the method proposed in this paper. 

\begin{figure}
    \centering
    \includegraphics[width=\columnwidth]{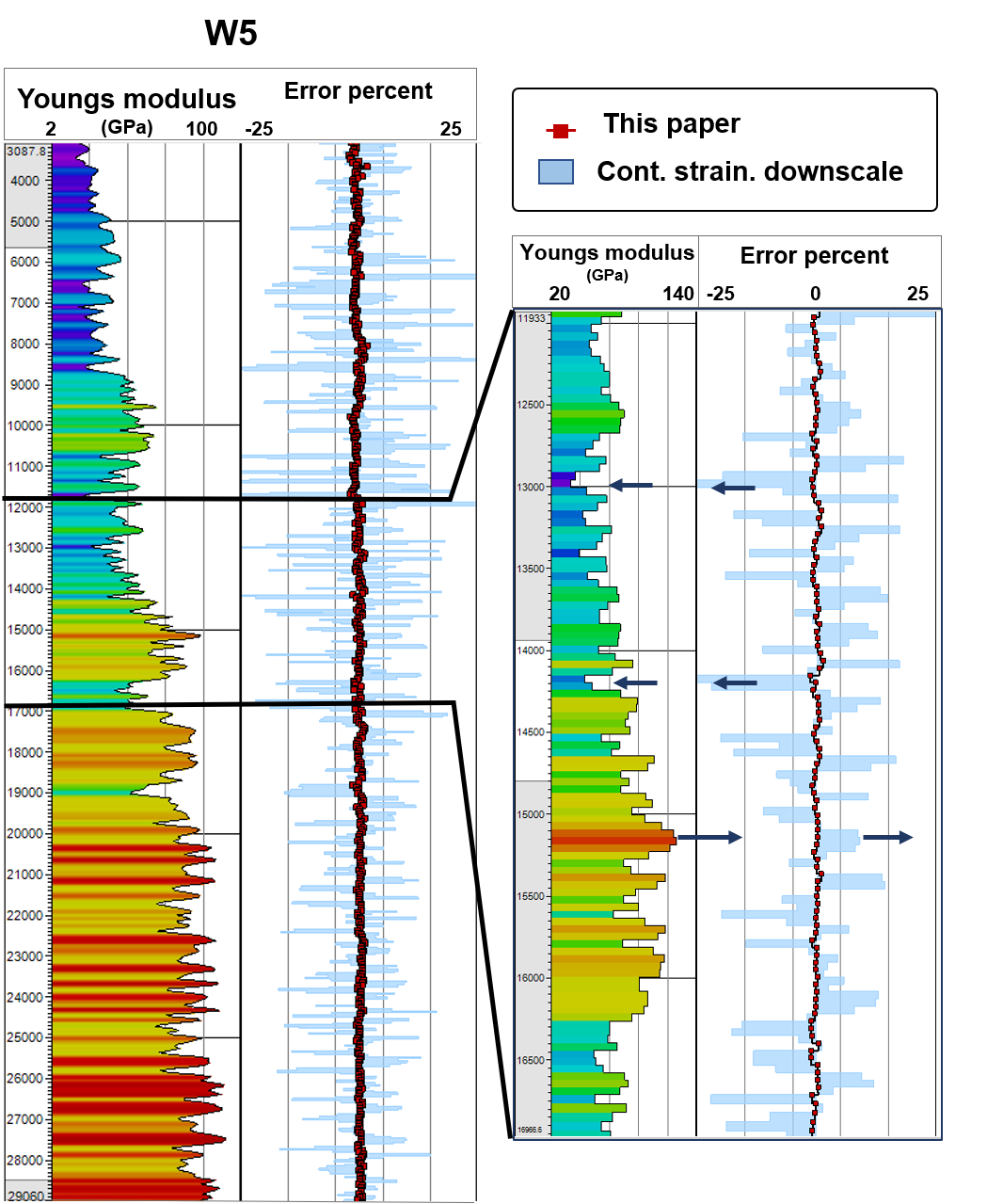}
    \caption{\scriptsize{Comparison between the constant strain downscale method \citep{herrera} and the ML stress prediction proposed in this paper.
    }\label{straindowscale2}}
\end{figure}

The two methods are further compared in Fig. \ref{straindowscale2} along one example vertical line near the center of the model. The figure shows that the error in the constant-strain downscale method is strongly correlated to the rock's stiffness $E$. The higher/lower the stiffness, the more the fracture gradient is over/under-estimated. In this example, the magnitude of the percent error can be 25\%. In comparison, the results obtained with the method proposed in this paper the fracture gradient is estimated within an error less than 1.3\% (0.33MPa), at all depths. 

\section{Conclusions}
The paper presented an upscale/downscale methodology to estimate the principal stress components in 3D models of the sub-surface. The results indicate that when departing from a coarse-resolution finite element model and a partial finite element solution obtained in a sub-volume of a high-resolution model, the method is able to approximate the fine detail of the solution that would be obtained in high-resolution in the entire 3D model. The results indicate that the estimates of the principal stress components are within a 2\% error in comparison to the finite element solutions at high resolution in an example case. 

The results indicate that the accuracy of the stress estimates is only weakly dependent on the magnitude and the local fluctuations of the stress field. Percentwise, the accuracy of the stress estimates was also observed to improve with depth. 

\section{Future directions}
One model was tested out of the infinite number of models that could have been tested. Not every scenario can be studied as an individual case, but one could consider groups of scenarios and few examples of each can be studied. In doing so, the limits of applicability of the method can be established. One group, for instance, could be scenarios with minor structural details where the mechanical properties are the main drivers for stress heterogeneity. The case treated here would fall in that category. Another group could include cases where local structures, such as small faults are the main variable affecting the stress field. Yet another group could include cases with regional stress perturbations due to for instance, salt bodies. This is currently a work in progress.
 
Tectonics is a main variable to consider. There are three tectonic regimes (normal, strike-slip and inverse) and we considered only one. We would expect that the higher the tectonic forces and anisotropy, the stronger the long-range dependency on the stress solutions. Some experiments that were recently carried out indicated that the method presented here would perform well in scenarios of strong tectonic forces. Yet, this is a work in progress and results will be presented in a forthcoming paper. 

There are several potential applications to the method proposed. One would be in sensitivity analysis. As of today, if one wanted to carry out a sensitivity analysis in relation to pressure, or rock stiffness for instance, many high-resolution finite element models would need to be simulated. This is extremely costly in time and computational resources. If a trained neural network model could serve as a proxy for those finite-element models, it would be of great practical interest. This is currently being investigated.

\section*{Acknowledgments}
This work was done in cooperation with the geomechanics experts in Schlumberger Center of 
Excellence in The UK. They provided outstanding advice on finite element modelling and geomechanics in general.
%\vspace{0 pt}

%%%%%%%%%%%%%%%%%%%% REFERENCES %%%%%%%%%%%%%%%%%%
%\newpage
\bibliography{deepdownscale}

\begin{thebibliography}{}
\expandafter\ifx\csname natexlab\endcsname\relax\def\natexlab#1{#1}\fi
\providecommand{\url}[1]{\href{#1}{#1}}
\providecommand{\dodoi}[1]{doi:~\href{http://doi.org/#1}{\nolinkurl{#1}}}
\providecommand{\doeprint}[1]{\href{http://ascl.net/#1}{\nolinkurl{http://ascl.net/#1}}}
\providecommand{\doarXiv}[1]{\href{https://arxiv.org/abs/#1}{\nolinkurl{https://arxiv.org/abs/#1}}}

\bibitem[{Abadi {et~al.}(2015)Abadi, Agarwal, Barham, Brevdo, Chen, Citro,
  Corrado, Davis, Dean, Devin, Ghemawat, Goodfellow, Harp, Irving, Isard, Jia,
  Jozefowicz, Kaiser, Kudlur, Levenberg, Man\'{e}, Monga, Moore, Murray, Olah,
  Schuster, Shlens, Steiner, Sutskever, Talwar, Tucker, Vanhoucke, Vasudevan,
  Vi\'{e}gas, Vinyals, Warden, Wattenberg, Wicke, Yu, \&
  Zheng}]{tensorflow2015-whitepaper}
Abadi, M., Agarwal, A., Barham, P., {et~al.} 2015, {TensorFlow}: Large-Scale
  Machine Learning on Heterogeneous Systems.
\newblock \url{http://tensorflow.org/}

\bibitem[{{Amadei} \& {Stephansson}(1997)}]{amadei}
{Amadei}, B., \& {Stephansson}, O. 1997, Rock Stress and Its Measurement, 1st
  edn. (Springer Netherlands), \dodoi{10.1007/978-94-011-5346-1}

\bibitem[{{Atienza}(2018)}]{atienza}
{Atienza}, R. 2018, Advanced Deep Learning with Keras: Apply Deep Learning
  Techniques, Autoencoders, GANs, Variational Autoencoders, Deep Reinforcement
  Learning, Policy Gradients, and More (Packt Publishing)

\bibitem[{Babaei \& King(2012)}]{babei}
Babaei, M., \& King, P.~R. 2012, Transport in Porous Media, 93, 753,
  \dodoi{10.1007/s11242-012-9981-4}

\bibitem[{{Backus}(1962)}]{backus}
{Backus}, George., E. 1962, Journal of Geophysical Research, 67(11), 4427,
  \dodoi{10.1029/JZ067i011p04427}

\bibitem[{Berard {et~al.}(2015)Berard, Desroches, Yang, Weng, \&
  Olson}]{Berard}
Berard, T., Desroches, J., Yang, Y., Weng, X., \& Olson, K.~E. 2015, in
  High-Resolution 3D Structural Geomechanics Modeling for Hydraulic Fracturing
  (Society of Petroleum Engineers. Hydraulic fracturing conference, The
  Woodlands, Texas, USA), \dodoi{10.2118/173362-MS}

\bibitem[{{Brouwer} \& {Fokker}(2013)}]{brouwer2013a}
{Brouwer}, G., \& {Fokker}, A., P. 2013, in {Upscaling and Downscaling with an
  Effective Medium Theory, Applied to Heterogeneous Reservoir} (Society of
  Petroleum Engineers. EAGE Annual Conference and Exhibition, London, UK),
  \dodoi{https://doi.org/10.2118/164803-MS}

\bibitem[{{Buck} {et~al.}(2013){Buck}, {Iliev}, \& {Andra}}]{buck}
{Buck}, M., {Iliev}, O., \& {Andra}, H. 2013, Open Mathematics, 11, 680 ,
  \dodoi{https://doi.org/10.2478/s11533-012-0166-8}

\bibitem[{{Castelletto} {et~al.}(2017){Castelletto}, {Hajibeygi}, \&
  {Tchelepi}}]{castelletto}
{Castelletto}, N., {Hajibeygi}, H., \& {Tchelepi}, H.~A. 2017, J. Comput.
  Phys., 331, 337–356, \dodoi{10.1016/j.jcp.2016.11.044}

\bibitem[{{Chalon} {et~al.}(2004){Chalon}, {Mainguy}, {Longuemare}, \&
  {Lemonnier}}]{chalon2004}
{Chalon}, F., {Mainguy}, M., {Longuemare}, P., \& {Lemonnier}, P. 2004,
  International journal for numerical and anlaytical methods in geomechanics,
  28(11), 1105, \dodoi{https://doi.org/10.1002/nag.379}

\bibitem[{{Chollet}(2017)}]{Chollet2017}
{Chollet}, F. 2017, Deep Learning with Python, 1st edn. (USA: Manning
  Publications Co.)

\bibitem[{{Chollet}(2015)}]{Chollet2015}
{Chollet}, F. e.~a. 2015, Keras,  GitHub.
\newblock \url{https://github.com/fchollet/keras}

\bibitem[{{Efendiev} {et~al.}(2013){Efendiev}, {Galvis}, \&
  {Hou}}]{Efendiev2013Book}
{Efendiev}, Y., {Galvis}, J., \& {Hou}, Thomas, Y. 2013, Journal of
  Computational Physics, 251, 116,
  \dodoi{https://doi.org/10.1016/j.jcp.2013.04.045}

\bibitem[{{Efendiev} \& {Hou}(2009)}]{Efendiev2009}
{Efendiev}, Y., \& {Hou}, T. 2009, {Applications of multiscale finite element
  methods. In: Multiscale Finite Element Methods. Surveys and Tutorials in the
  Applied Mathematical Sciences}, Vol.~4 (Springer, New York, NY), 1--69,
  \dodoi{https://doi.org/10.1007/978-0-387-09496-0\_5}

\bibitem[{{Fjaer} {et~al.}(1992){Fjaer}, {Horsrud}, {Raaen}, {Risnes}, \&
  {Holt, R. M.}}]{fjaer}
{Fjaer}, E., {Horsrud}, P., {Raaen}, A.~M., {Risnes}, R., \& {Holt, R. M.}
  1992, Petroleum related rock mechanics, Vol.~33 (Elsevier)

\bibitem[{Fl{\'o}rio \& Almeida(2015)}]{florio}
Fl{\'o}rio, R., \& Almeida, J. 2015, in A comparative study of the tensor and
  upscaling methods for evaluating permeability in fractured reservoirs
  (International Association for Mathematical Geology (IAMG), 17th Annual
  Conference of the International Association for Mathematical Geosciences),
  465--474, \dodoi{10.13140/RG.2.1.1433.3922}

\bibitem[{{Garcia} {et~al.}(2013){Garcia}, {Nagel}, {Zhang}, \& {Lee}}]{garcia}
{Garcia}, X., {Nagel}, N., {Zhang}, F., \& {Lee}, B. 2013, in {Revisiting
  Vertical Hydraulic Fracture Propagation Through Layered Formations - A
  Numerical Evaluation. ARMA 13-203}, Vol.~4 (American Rock Mechanics
  Association. 47th US Rock Mechanics and Geomechanics Symposium), 2501--2511.
\newblock
  \url{http://www.scopus.com/inward/record.url?eid=2-s2.0-84892878236\&partnerID=tZOtx3y1}

\bibitem[{Gaur \& Simonovic(2019)}]{gaur}
Gaur, A., \& Simonovic, S.~P. 2019, in Trends and Changes in Hydroclimatic
  Variables, ed. R.~Teegavarapu (Elsevier), 199 -- 273,
  \dodoi{https://doi.org/10.1016/B978-0-12-810985-4.00004-9}

\bibitem[{Goodfellow {et~al.}(2016)Goodfellow, Bengio, \&
  Courville}]{Goodfellow-et-al-2016}
Goodfellow, I., Bengio, Y., \& Courville, A. 2016, Deep Learning (MIT Press).
\newblock \url{http://www.deeplearningbook.org}

\bibitem[{Gulli \& Pal(2017)}]{gulli2017deep}
Gulli, A., \& Pal, S. 2017, Deep Learning with Keras (Packt Publishing).
\newblock \url{https://books.google.com/books?id=20EwDwAAQBAJ}

\bibitem[{Ita \& Malekzadeh(2015)}]{ita2015a}
Ita, J., \& Malekzadeh, F. 2015, A True Poroelastic Up and Downscaling Scheme
  for Multi-scale Coupled Simulation (Society of Petroleum Engineering,
  Reservoir Simulation Symposium, 23-25 February, Houston, Texas.
  SPE-173253-MS), \dodoi{https://doi.org/10.2118/173253-MS}

\bibitem[{{Koupriantchik} {et~al.}(2005){Koupriantchik}, {Hunt}, \&
  {Meyers}}]{koupriantchik}
{Koupriantchik}, D., {Hunt}, S. P.~R., \& {Meyers}, A.~G. 2005, in
  Geomechanical Modeling of Salt Diapirs: A Field Scale Analysis For a 3D Salt
  Structure From the North Sea (International Society for Rock Mechanics and
  Rock Engineering. EUROCK, 18-20 May, Brno, Czech Republic),
  \dodoi{https://doi.org/10.2118/93605-MS}

\bibitem[{Maraun \& Widmann(2018)}]{maraunwidmann2018}
Maraun, D., \& Widmann, M. 2018, Statistical Downscaling and Bias Correction
  for Climate Research (Cambridge University Press),
  \dodoi{10.1017/9781107588783}

\bibitem[{Maraun {et~al.}(2010)Maraun, Wetterhall, Ireson, Chandler, Kendon,
  Widmann, Brienen, Rust, Sauter, Themeßl, Venema, Chun, Goodess, Jones, Onof,
  Vrac, \& Thiele-Eich}]{maraun2010}
Maraun, D., Wetterhall, F., Ireson, A.~M., {et~al.} 2010, Reviews of
  Geophysics, 48(3), \dodoi{10.1029/2009RG000314}

\bibitem[{{Menezes} \& {Gosselin}(2003)}]{menezes2003}
{Menezes}, C., A., \& {Gosselin}, R., O. 2003, From Logs Scale to Reservoir
  Scale: Upscaling of the Petro-Elastic Model (Society of Petroleum Engineers,
  Europec/EAGE Annual Conference and Exhibition, 12 - 15 June, Vienna,
  Austria), \dodoi{https://doi.org/10.2118/100233-MS}

\bibitem[{{Nunna} \& {King}(2017)}]{nunna2017a}
{Nunna}, K., \& {King}, Michael, J. 2017, in {Dynamic Downscaling and Upscaling
  in High Contrast Systems} (Society of Petroleum Engineers, Reservoir
  Simulation Conference, 20-22 February,Montgomery, Texas. SPE-182689-MS),
  \dodoi{https://doi.org/10.2118/182689-MS}

\bibitem[{{Qiu} {et~al.}(2005){Qiu}, {Marsden}, {Solovyov}, {Safdar},
  {Chardac}, \& {Shatwan}}]{qiu}
{Qiu}, K., {Marsden}, R., {Solovyov}, Y., {et~al.} 2005, in Downscaling
  Geomechanics Data for Thin-Beds Using Petrophysical Techniques, SPE-93605-MS
  (Society of Petroleum Engineers. Middle East Oil and Gas Show and Conference,
  Bahrain, Kingdom of Bahrain), \dodoi{https://doi.org/10.2118/93605-MS}

\bibitem[{{Ren} {et~al.}(2013){Ren}, {McLennan}, {Cunha}, \& {Deutsch}}]{ren}
{Ren}, W., {McLennan}, J., {Cunha}, Luciane, B., \& {Deutsch}, C.~V. 2013, in
  An Exact Downscaling Methodology in Presence of Heterogeneity: Application to
  the Athabasca Oilsands (Society of Petroleum Engineers. International Thermal
  Operations and Heavy Oil Symposium, 1-3 November, Calgary, Canada),
  \dodoi{https://doi.org/10.2118/97874-MS}

\bibitem[{{Rodriguez-Herrera} {et~al.}(2020){Rodriguez-Herrera},
  {Garcia-Teijeiro}, {Herwanger}, {Koutsabeloulis}, Tone, \& Jalal}]{herrera}
{Rodriguez-Herrera}, A., {Garcia-Teijeiro}, X., {Herwanger}, J., V., {et~al.}
  2020, Systems and methods for downscaling stress for seismic-driven
  stochastic geomechanical models. U.S patent number US20150112656A1.
\newblock \url{https://patents.google.com/patent/US20150112656A1/en}

\bibitem[{Sailor {et~al.}(2000)Sailor, Hu, Li, \& Rosen}]{sailor2000}
Sailor, D., Hu, T., Li, X., \& Rosen, J. 2000, Renewable Energy, 19(3), 359,
  \dodoi{https://doi.org/10.1016/S0960-1481(99)00056-7}

\bibitem[{{Smith}(2020)}]{SmithOnline}
{Smith}, G. 2020, Density\/ logging,  Petro-wiki.
\newblock \url{https://petrowiki.org/Density\_logging}

\bibitem[{{Tarmizi} \& {Hatin}(2019)}]{tarmizi}
{Tarmizi}, A., \& {Hatin}, A. 2019, International Journal of Integrated
  Engineering, 11(1).
\newblock
  \url{https://publisher.uthm.edu.my/ojs/index.php/ijie/article/view/4288}

\bibitem[{{Timosenko} \& {Gere}(2012)}]{timosenko}
{Timosenko}, S., \& {Gere}, James, M. 2012, Theory of elasticity, 3rd edn.,
  Engineering societies monographs (Mcgrw-Hill Education (India), Pvt Limited)

\bibitem[{{Torrealba} {et~al.}(2019){Torrealba}, {Hoteit}, \&
  {Chawathe}}]{torrealba2019a}
{Torrealba}, A., {Hoteit}, H., \& {Chawathe}, A. 2019, SPE Reservoir Evaluation
  and Engineering, 22(4), \dodoi{https://doi.org/10.2118/1a87276-PA}

\bibitem[{{Trudeng} {et~al.}(2014){Trudeng}, {Garcia-Teijeiro},
  {Rodriguez-Herrera}, \& {Khazanehdari}}]{trudeng}
{Trudeng}, T., {Garcia-Teijeiro}, X., {Rodriguez-Herrera}, A., \&
  {Khazanehdari}, J. 2014, in Using Stochastic Seismic Inversion as Input for
  3D Geomechanical Models (European Association of Geoscientists and Engineers.
  IPTC International Petroleum Technology Conference),
  \dodoi{https://doi.org/10.3997/2214-4609-pdb.395.IPTC-17547-MS}

\bibitem[{{Ueda}(2018)}]{Ueda}
{Ueda}, K. e.~a. 2018, in { Hydraulic Fracture Design in the Presence of
  Highly-Stressed Layers: A Case Study of Stress Interference in a
  Multi-Horizontal Well Pad, 189845-MS SPE} (Society of petroleum engineeing.
  Hydraulic fracturing conference, The Woodlands, Texas.),
  \dodoi{https://doi.org/10.2118/189845-MS}

\bibitem[{Vandal {et~al.}(2019)Vandal, Kodra, \& Ganguly}]{vandal}
Vandal, T., Kodra, E., \& Ganguly, A.~R. 2019, Theoretical and Applied
  Climatology, 137, 557, \dodoi{10.1007/s00704-018-2613-3}

\bibitem[{{Warpinski} {et~al.}(1982){Warpinski}, {Schmidt}, \&
  {Northrop}}]{warpinski}
{Warpinski}, R., N., {Schmidt}, A., R., \& {Northrop}, A., D. 1982, Journal of
  Petroleum Technology, ARMA 13-203, 653, \dodoi{doi:10.2118/8932-PA}

\bibitem[{{Xu}(1999)}]{xu}
{Xu}, C.-y. 1999, Progress in Physical Geography: Earth and Environment, 23
  (2), 229, \dodoi{10.1177/030913339902300204}

\bibitem[{{Zoback}(2007)}]{zoback2007}
{Zoback}, M.~D. 2007, Reservoir Geomechanics (Cambridge University Press),
  \dodoi{10.1017/CBO9780511586477}

\end{thebibliography}
\bibliographystyle{aasjournal}
%\bibliographystyle{unsrt}

%%% Figure Sets
%\include{figure_sets}

\end{document}